# Ebola Optimization Search Algorithm: A new metaheuristic algorithm based on the propagation model of Ebola virus disease

Olaide N. Oyelade[1] and Absalom E. Ezugwu[1]

[1]School of Mathematics, Statistics, and Computer Science, University of KwaZulu-Natal, King Edward Avenue, Pietermaritzburg Campus, Pietermaritzburg, 3201, KwaZulu-Natal, South Africa

Corresponding authors: Absalom E. Ezugwu (email: ezugwua@ukzn.ac.za).

**Abstract.** Ebola, also known as Ebola virus disease or Ebola hemorrhagic fever, is a viral hemorrhagic fever of humans and other primates caused by ebolaviruses. The Ebola virus and the disease in effect tend to randomly move individuals in the population around susceptible, infected, quarantined, hospitalized, recovered, and dead sub-population. Motivated by the effectiveness in propagating the disease through the virus, a new bio-inspired and population-based optimization algorithm is proposed. This paper presents a novel metaheuristic algorithm named Ebola optimization algorithm (EOSA). To correctly achieve this, this study models the propagation mechanism of the Ebola virus disease, emphasising all consistent states of the propagation. The model was further represented using a mathematical model based on first-order differential equations. After that, the combined propagation and mathematical models were adapted for developing the new metaheuristic algorithm. To evaluate the proposed method's performance and capability compared with other optimization methods, the underlying propagation and mathematical models were first investigated to determine how they successfully simulate the EVD. Furthermore, two sets of benchmark functions consisting of forty-seven (47) classical and over thirty (30) constrained IEEE CEC-2017 benchmark functions are investigated numerically. The results indicate that the performance of the proposed algorithm is competitive with other state-of-the-art optimization methods based on scalability, convergence, and sensitivity analyses. Extensive simulation results indicate that the EOSA outperforms other state-of-the-art popular metaheuristic optimization algorithms such as the Particle Swarm Optimization Algorithm, Genetic Algorithm, and Artificial Bee Colony Algorithm on some shifted, high dimensional and large search range problems.

1. Introduction

Ebola virus represents the virus causing the Ebola virus disease (EVD). The disease was first so named in the Democratic Congo Republic (DRC) in 1976. A widespread catastrophic outbreak was reported in late 2013 in the West African regions, including Sierra Leone, Liberia, Mali, Nigeria, and Senegal. It is widely reported that the virus made its entry into the human population through consumption or contact with infected animals such as fruit bat (Gumusova, Sunbul, & Leblebicioglu, 2015), (Osterholm, et al., 2015), (Kadanali & Karagoz, 2015). This animal-to-human infection led to person-to-person, hence becoming an epidemic across the West African region. Contrary to the novel corona virus (COVID-19), the EVD person-to-person transmission occurs only when the infected person exhibits some form of signs and symptoms associated with Ebola. This transmission is aided by contact with any form of body fluid of an infected person, and also when a healthy person comes in contact with infected objects since the Ebola virus can survive on dry surfaces, like doorknobs and countertops, for several hours (CDC, 2021), (Tanade, Pate, Paljug, Hoffman, & Wang, 2019). The haemorrhagic disease, known to be notoriously fatal has been reported to have mortality rates ranging from 25% to 90%. An average of 50% isnly due to fluid loss rather than blood loss (The Agent, 2020), (WHO, 2020). Although the experimental Ebola vaccine proved highly protective against EVD, the transmission rate from the infected to the susceptible population is alarming. The high survival rate of EBOV in body fluids, including breast milk, saliva, urine, semen, cerebrospinal fluid, and aqueous humor, in addition to blood and blood derivatives, and detected in amniotic fluid, tears, skin swabs and stool by reverse transcription (RT)-PCR, presents a very high infection and transmission rate. This then implies that a one-time entry of the virus into a susceptible population through a single individual has a high propagation rate among the population.

A closer study of the propagation strategy of the EVD and the resulting propagation model inspired the metaheuristic algorithm proposed in this study. Deriving computational solution from natural phenomenons has promoted a field of computing referred to as nature-inspired computing. A broader view of this aspect of computing may well relate with the field of artificial intelligence (AI) and also computational intelligence (CI), where computational systems are designed by synthesizing behaviors of organisms or natural phenomenon (Marrow, 2000), (Wang, et al., 2017), (Siddique, 2015). Metaheuristic algorithms are nature-inspired optimization solutions with high performance and lower required computing capacity which has successfully solved complex real-life problems in Engineering, medical sciences and sciences, especially in areas concerning swarm intelligence based algorithms (XS, 2013), (Green, Aleti, & Garcia, 2017), (SwatiSwayamsiddha, 2020). These optimization algorithms are designed without specific to a particular problem and are often categorized by the ability to perform a local or global search, handle single-solution or whole population, use memory, adopt greedy or iterative search process. The techniques often achieve near-optimal solutions to large-scale optimization problems due to their highly flexible manner of operation and ability to learn quickly owing to their natural or biological systems from which their designs were inspired.

A subfield of natural computation consists of biology-inspired techniques which also referred to as bio-inspired algorithms or computational biology. These techniques are stochastic in nature, far from the design of deterministic heuristics (Ezugwu, Adeleke, Akinyelu, & Viriri, 2020). This feature has made it possible to represent the biological evolution of nature, hence capable of being used as a global optimization solution. Recently, bio-inspired optimization algorithms have helped support machine learning to address the optimal solutions of complex problems in science and engineering (Darwish, 2018). The bio-inspired algorithms combine biological concepts with mathematics and computer sciences and are broadly classified as evolutionary algorithms (EA), biology algorithms, and swarm intelligence (SI) algorithms. Although the last two categories are often combined and referred to as swarm intelligence, we found that not all bio-inspired algorithms have the swarm feature. Examples of the evolutionary algorithms are Genetic algorithms (GA) (Sivanandam & Deepa, 2008), Genetic programming (GP), differential evolution (DE), the evolution strategy (ES), Coral Reefs Optimization Algorithm (CRO) (Ezugwu, et al., 2021) and evolutionary programming (EP). Examples of SI-based algorithms are Food foraging behavior of honeybees Artificial Bee Colony (ABC), Echolocation ability Ant-lion optimizer (ALO), Luciferin induced glowing behavior Bees Algorithm (BAO), Bat Algorithm (BOA), ), Hunting behavior Barnacles mating optimizer (BMO), Swarming around hive by honey bees Cuckoo optimization algorithm (COA), Echo-cancellation Cuckoo search (CSO), Hunting behavior & social hierarchy Dolphin echolocation (DEO), Social interaction and food foraging Dragonfly algorithm (DFA), Static and dynamic swarming behavior Deer hunting optimization (DHO), Pollination process of flowers Fire-fly algorithm (FFA), Food foraging behavior Hunting search (FFO), Bubble-net hunting Fruit fly optimization (FOA), Obligate brood parasitic behavior Flower-pollination algorithm (FPA), Navigation and foraging behaviors Grasshopper optimization algorithm (GOA), Spiral flying path of moth Glowworm Swarm Optimization (GSO), Cuckoos' survival efforts Grey wolf optimizer (GWO), Flashing light patterns Moth-flame optimization (MFO), Mating behavior Manta ray foraging optimizer (MRFO), Hunting behavior of humans SailFish optimizer (SFO), Group hunting behavior Salp swarm algorithm (SSA), and Hunting mechanism Whale optimization algorithm (WOA).

These EA and SI based algorithms have demonstrated good performance at solving real-world complex combinatorial problems, which are considered to be a fundamentally vital and critical task (Agushaka & Ezugwu, 2020), (Munien, et al., 2020), (Munien & Ezugwu, Metaheuristic algorithms for one-dimensional bin-packing problems: A survey of recent advances and applications, 2021), (Achary, et al., 2021), (Ezugwu, Adewumi, & & Frîncu, 2017), (Ezugwu A. E., Enhanced symbiotic organisms search algorithm for unrelated parallel machines manufacturing scheduling with setup times, 2019), (Ezugwu & Prayogo, Symbiotic organisms search algorithm: theory, recent advances and applications, 2019). In addition, studies have shown their capability to efficiently scale up to handle large scale problems as opposed to traditional optimization methods, which are more effective for small-scale problems (Game, Vaze, & Emmanuel, 2020). Further research in bio-inspired computing areas will lead to achieving similar and better new optimization algorithms capable of solving modern-day optimization problems. Our study showed that exploring the propagation model of diseases with an endemic and pandemic nature may yield an outperforming optimization algorithm with interesting performance in solving real-world optimization problems. This study considers that the exploration and exploitation phases of optimization algorithms are practically coupled into the natural order and strategy of propagating these diseases. The study confirms that finding a good balance between exploitation and exploration of the problem search space for an optimization algorithm determines its ability to find a globally optimal solution (CREPINSEK, LIU, & MERNIK, 2012), (Cuevas, Echavarría, & Ramírez-Ortegón, 2014). The exploration phase often allows for finding candidate solutions that are not neighbor to the current solution, while exploitation maintains its search in the neighborhood. Hence, we found a balance of the two scenarios in the disease propagation model for escape from a local optimum with no neglect to good solutions in the neighborhood.

This study is motivated by the propagation strategy of the Ebola virus disease, and in general, the capability of bio-inspired algorithms to provide abundant inspiration for the design and implementation of intelligent algorithms that are powerful tools for solving real-life problems. We propose a novel metaheuristic algorithm, which is an optimization algorithm inspired by the Ebola virus disease and its propagation model. The proposed algorithm is referred to as the Ebola Optimization search algorithm (EOSA). The main advantage of this algorithm over similar optimization algorithms is that it presents a dynamic mechanism for profitably updating the population as they transit through susceptible, infection, quarantine, recovered, and hospitalized operations for a better fit. To quantitatively measure how fit a given solution is in solving the problem, giving intuitive results, discovering the best or worst candidate solution, the resulting optimization algorithm is investigated on about fifty (50) classical benchmark optimization functions (Jamil & Yang, 2013) and more than thirty (30) CEC functions (Cheng, et al., 2018). In summary, the main contributions of this research are as follows:

i. A new nature-inspired metaheuristic algorithm called EOSA inspired by the propagation model of the Ebola virus disease is proposed.
ii. Several experiments are conducted using over 89 mathematical optimization problems, including the classical benchmark functions and CEC'17 test suite, which are considered to be challenging test problems in the literature to evaluate the efficiency of the proposed EOSA. The results of these experiments may serve as important inputs for further research.
iii. Validation of the obtained numerical results using statistical analysis test, which further support the superiority claim of the proposed EOSA optimization method over the existing state-of-the-art metaheuristic algorithms.

The rest of the paper is organized as follows. Section 2 provides an overview of the Ebola virus dieses. The proposed propagation model, mathematical model, and algorithmic design for the EOSA algorithm are given in Section 3. Sections 4 details the benchmark functions applied to evaluate the performance of the proposed algorithm. Also, the section listed and discussed the parameterization and assignment of initial values used for experimentation. A discussion on results obtained is presented in Section 5, including numerical simulations that support the proposed propagation model. A detailed comparative analysis of the performance of EOSA and similar algorithms are also presented in the section. In Section 6, we give concluding remarks on how our novel optimization algorithm fit in the literature, its real-life applicability, and perspectives for future works.

2. **Overview of Ebola and Related Work**

This section presents a summary of the Ebola virus disease, its propagation technique, and relevant SIR-based models that support the modeling presented in this study. Also, considering the nature of the optimization algorithm proposed in this study, which shares some biology principles, we review studies that have developed bio-inspired optimization algorithms. The phenomenon behind the resulting algorithms is briefly stated and reported performance or application of the algorithms is also expatiated.

**2.1 The Ebola Virus (EBOV) and Ebola Virus Diseases (EVD): The Propagation Mechanism**

Ebola viruses result in what is known as the Ebola virus disease (EVD) once they successfully infect the host in a manner suggesting victimization of the host. They are classified among the family of Filoviridae viruses, which are recognized by their different shapes of short or elongated branched filaments sizing up to 14,000 nanometers in length (The Agent, 2020). About six different species of the EBOV have been reported to exist. These are the Bundibugyo Ebola virus, Ebola-Zaire virus, Tai Forest Ebola virus, and Sudan Ebola virus account for large flare-ups or outbreaks in Africa.

Exposure of a human individual to the virus through pathogenic agents or contaminated environment initiates a population-based infection and thereafter propels the spread or propagation of the disease. Direct contact with infecteded individual spur the propagation and spread of the virus. This contact relies on broken skin or mucous membranes in the eyes, nose, mouth, or other openings. It is assumed that such openings in the human body allow for body fluids (e.g. urine, saliva, sweat, faeces, vomit, breast milk, amniotic fluid, blood and semen) bearing the virus to be transmitted to other susceptible individuals. Another host to the Ebola virus, which may transmit the disease to a healthy or susceptible individual is a contaminated environment. An environment, such as medical equipment's, clothes, beddings and other related utensils, is considered to be contaminated if the body fluid of an infected individual has been spilt within such environment or object. Whereas an infected individual and a contaminated environment appear to have enhanced the propagation of EVD, infected animals consumed by humans have also shown to propagate the disease (Rewar & Mirdha, 2014). These animals include bats, chimpanzees, fruit bats, and forest antelope, which are often hunted for food. Another propagative mechanism of the EBOV culturally driven is burial practices in the most affected population and regions. This is mainly transmitted through contact with infected dead bodies. Ebola virus is not propagated through the air.

Different strategies ranging from case-based management approach, surveillance and contact tracing, quarantine of infected cases, infection prevention and control practices, and safe burial rites have been adopted in reviving and surviving infected cases. However, infected cases remain positive while the virus remains in their blood. The EBOV infection and propagation rate presents an appealing computational solution to numerous problems and so motivated the design of the proposed metaheuristic algorithm. While it appears that the solutions for mitigating the spread of the virus are suggestive of scaling down the infection rate, we argue that some other factors are still contributory to the propagation model. For instance, it is widely reported that the time-scale from symptom onset to death is an average of 10 days in 50–90% of cases (Mobula, et al., 2018).

To formalize and apply the propagation model of EBOV, we review some susceptible-infection-recovery (SIR) models. This is necessary for mainstreaming the concept proposed in the study. An interesting SIR model, based on EBOV, combining agent-based and compartmental model, has been presented (Tanade, Pate, Paljug, Hoffman, & Wang, 2019). The authors suggested that the hybrid model can switch from one paradigm to another on a stochastic threshold. The agent-based model consists of Susceptible (S), Infected (I), Hospitalized (H), Recovered (R), Funeral (F), and Dead (D).  In the compartmental model, the Exposed (E) item was added to make up seven compartments. The SIR-based model was proposed to model the movement of individuals in a population from one compartment to another in both paradigms. For instance, individuals may move from Susceptible (S), Infected (I), Hospitalized (H), based on pre-existing computed rates. One external compartment considered in the literature is the influence of EBOV carrying animals like bats. The assumption made was that since these animals can infect the human population without them (the animals) becoming ill, they present a reservoir-like mechanism for the virus in the SIR model. Furthermore, the authors assumed that rate of infection and hospitalization between infected individuals who will recover or die is the same, deceased individual is buried in unsafe practices, and that recovered individual are removed from the system. This SIR model presents a foundation for the modeling and implementation of the optimization algorithm proposed in this study. We considered that the compartments defined by the work of Tanade *et al.* demonstrates the possibility to monitor and simulating the EBOV propagation model for the optimization task in our study.

In related work, Berge *et al,* also modeled the propagation model of EBOV using the SIR-type model. The study's novelty was the addition of the role of the indirect environmental transmission on the dynamics of EVD and to assess the effect of such a feature on the long run of the disease (Berge, Lubuma, Moremedi, Morris, & Kondera-Shava, 2017). The authors showed that factoring direct and indirect transmission of EBOV into a SIR model promotes a system where the virus always exists in a population, increasing the propagation rate. Taking a cue from the novelty of this work in addition to that of Tanade *et al.,* we adapt the model proposed in this study to support the concept of direct and indirect transmission promoted by Berge *et al.* both studies supported their SIR models with mathematical models and further simulation for validating the performance of their model. Similarly, Yet (2019) successfully represented the basic interactions between EBOV and wild-type Vero cells in vitro (Yet, 2019). Rafiq et al. also proposed the SEIR model, which the mathematical model supported demonstrating the dynamics and illustrating the

Ebola virus's stability pattern in the human population. We found that their mathematical model which is of the form couple linear differential equation. The authors applied their SEIR model to study the disease-free equilibrium (DFE) and endemic equilibrium (EE) and thereafter report on the stability of the model. In another study that investigated the spread of EVD in India, the authors (Sau, 2017) investigated EBOV transmission in the region through an SEIR model they proposed. Using ordinary differential equations, the study represented the SEIR model as a mathematical model and further simulated it using a spatiotemporal epidemiology modeler (STEM). Rachah and Torres also applied a mathematical model for the study of the outbreak of EBOV and eventually the EVD (Rachah & Torres, 2015). The novelty of this study is the addition of the use of vaccination to the proposed model. We found this appealing considering the role of the vaccine in stemming the tide of the infected population. Whereas most SIER approaches have often adopt the stochastic method for the simulation of the model, Okyere et al., considered the use of a deterministic scheme for designing models and studying the infection rate of EBOV (Okyere, Ankamah, Hunkpe, & Mensah, 2020). As an improvement to the work of Rachah and Torres, which factored in vaccination, the study also captured treatment and educational campaign as time-dependent control functions in the SEIR model proposed.

Considering the review presented above, this study developed an SEIR-based model which added more compartments. The proposed SEIR model factored in the notion of quarantine, which we found to play a role in curtailing EBOV propagation. In addition, we modeled the SEIR model to allow for the inclusion of the influence of vaccine in the pace of the growth of infection among a given population. The SEIR model was then formulated using an ordinary differential equation. This presented a good understanding of the design of the proposed metaheuristic algorithm. We found this necessary due to the importance of all the compartments of the SIER model in achieving a population that successfully translates between S-I. The resulting model is detailed in Section 3 and its supporting mathematical model. Having considered the biological perspective of the inspiration of the proposed algorithm, the following sub-section presents a review of some related metaheuristic algorithms motivated by nature-biology behavioral patterns.

## 2.2 Metaheuristic optimization algorithms: Bioinspired-Based Algorithms

Bio-inspired optimization algorithm represents a class of metaheuristic algorithms whose principles are inspired by a biology-nature phenomenon and have been successfully applied to solve different problems (Oliveira, Pires, Boaventura-Cunha, & Martins, 2020). This category of algorithms exploits the basic process of nature and then translates them into rules or procedures, which are then model computational for solving complex real-life problems. The mostly population-based algorithms and examples of such are Satin Bowerbird Optimizer (SBO), Earthworm Optimisation Algorithm (EOA), Wildebeest Herd Optimization (WHO), Virus Colony Search (VCS), Slime Mould Algorithm (SMA), Invasive weed colonization optimization (IWO), Biogeography-based optimization (BBO), Coronavirus optimization algorithm (COA), emperor penguin and salp swarm algorithm (ESA). Although evolutionary-based algorithms like GA and DE and swarm-based algorithms like PSO, WOA, and ABC share some characteristics of a biology-inspired algorithm, we have chosen to limit our review to those mentioned earlier.

ESA is a hybrid of two phenomena drawn from the Salp swarm algorithm and emperor penguin. The behaviour of the two creatures is modelled to achieve ESA. Comparing the proposed algorithm with similar metaheuristic algorithms, authors (Dhiman, 2019) revealed that the algorithm demonstrated good performance based on sensitivity, scalability, and convergence analyses. Coronavirus optimization algorithm (COA) based on its propagation strategy, and another variant namely Coronavirus herd immunity optimizer (CHIO) based on human immunity, have been proposed. The COA proposed in (Martínez-Álvarez, et al., 2020) and CHIO in (Al-Betar, Alyasseri, Awadallah, & Doush, 2020) leverages on herd infection and herd immunity, respectively. The effectiveness of COA was evaluated by applying it to the problem of the design of a convolutional neural network (CNN), while CHIO proved robust at real-world engineering problems. Earthworm Optimisation Algorithm (EOA), also referred to as EWA, is a metaheuristic algorithm whose inspiration was drawn from the reproductive nature of the earthworm (Wang, Deb, & Coelho, 2018). The mechanism involves two reproduction strategies where the first strategy allows for a parent to reproduce only one offspring while the other allows for more than one offspring. This reproducibility is controlled by the Caunchy mutation approach allowing for crossover operators.

Biogeography-Based Optimization (BBO) solves its optimization problem by implementing the concept for geographical distribution and positioning of the biological organism (Simon, 2008). Although BBO's features are similar to those of GAs, the authors drew inspiration from the original mathematical model of the biogeography of organisms to derive BBO. Experimentation shows that BBO successfully solved real-world sensor selection problems for detection of the status of aircraft engines and a selection of 14 benchmark optimization functions. Invasive Weed Optimization (IWO) is an optimization algorithm that has been widely applied to numerous problems and is based on a numerical stochastic optimization algorithm learned from the invasive nature of weeds (Mehrabian & Lucas, 2006). The aggressive invasive nature of weeds allows for colonizing the environment against other economically viable plants. Knowing that this is a disadvantage agricultural-wise, the concept has benefited the task of solving optimization problems. The resulting optimization algorithm was successfully applied to engineering problems, namely, optimizing and tuning the robust controller and well-known benchmark functions. Satin bowerbird optimization (SBO) is a biology-based optimization algorithm whose inspiration was drawn from the phenomenon of male satin bowerbird capable of attracting the female for breeding. (Zhang, Zhou, & Luo, 2019). The Satin Bowerbird Optimizer (SBO) optimization algorithm has been successfully applied to the problem of optimization in the estimation of efforts needed to develop software. Wildebeest Herd Optimization (WHO) is a bio-inspired metaheuristic algorithm rooted in wildebeest behaviour when searching for food (Amali, Bessie, & Dinakaran, 2019). The search is often guided by lookout for grazing land where there is a high density of vegetation. The WHO exploits the following natural characteristics of herds of wildebeest to achieve its performance: local search capability of wildebeest due to limited eyesight, look out for sparsely grazed region to avoid crowded grazing, exploitation of past experiences to explore regions with a high density of vegetation, starvation avoidance strategy deployed through a transition to new regions or location, and lastly, herd-based movement to avoid predators.

The propagation strategy of the virus in the host environment can sometimes be aggressive and often overwhelm the whole environment. Motivated by this mechanism, authors (DongLi, Zhao, Weng, & Han, 2016) proposed Virus Colony Search (VCS). The VCS exploration and exploitation phases leverage the propagation approach of the virus through diffusion or infection of the host environment. VCS has been successfully applied to the classic benchmark functions and the modern CEC2014 benchmark functions and real-life problem regarding energy consumption management (Jayasena, Li, Abd Elaziz, & Xiong, 2018). Slime Mould Algorithm (SMA) optimization algorithm is based on a fungus named slime mould inhabiting cold and humid places (Li, Chen, Wang, Heidari, & Mirjalili, 2020). The algorithm authors explored the nutritional stage, also referred to as plasmodium, of the organism for its design. They have a mechanism for multiple food sources at the same time to form a connected venous network so that they can even grow to more than 900 square centimetres depending on food availability. Using a mathematical model, authors were able to simulate the process of producing positive and negative feedback of the propagation wave of slime mould based on bio-oscillator to form the optimal path for connecting food with excellent exploratory ability and exploitation propensity. SMA was successfully applied to solve engineering problems, including cantilever, welded beam, and pressure vessel structure problems.

Although bio-inspired optimization algorithms are often classified into evolutionary and swarm intelligence-based approaches, we have presented a review of those we termed as biology-based. While we acknowledge that these are not exhaustive and many new optimization algorithms inspired by natural processes are developed day-by-day, they provide readers with a general understanding of the inspiration and principle behind such a class of algorithms.

## 3. EOSA: Methodology

Understanding how an SEIR-based model works in modeling the propagation of a disease is important to appeal to an optimization algorithm's design. Hence, this study proposes an improved SEIR-HDVQ model based on recent literature on EVD. Secondly, a presentation of the procedural flow of EOSA and the corresponding flow chart is designed and discussed. Lastly, to formalize the proposed optimisation algorithm, we represent it using a mathematical model and then design it. Therefore, this section is a detailed work on the methods applied to creating the new metaheuristic algorithm.

### 3.1 Model of EOSA

SEIR-based models designed for EVD have been proposed in the literature to monitor direct and indirect propagation of the disease in the selected population (Berge, Lubuma, Moremedi, Morris, & Kondera-Shava, 2017) and (Tanade, Pate, Paljug, Hoffman, & Wang, 2019). This study adopts and adapts two relevant models from the existing SEIR models by identifying and adding new compartments perceived to have been omitted. This study proposed adding the external agent or contaminated environment serving as a reservoir of the virus, vaccination, and quarantine, which PE, V represents, and Q, respectively. This became necessary considering that the Ebola virus and disease are not propagated among the human population except an individual is infected from the reservoir. Also, the roles played by vaccination and quarantining infected individuals have impacted the virus's propagation rate and disease. This perception is now reported in studies on the propagation of endemic or pandemic cases (Moghadas, et al., 2020), (BBC, 2021) (Mathebula, Ndwandwe, Pienaar, & Wiysonge, 2019), (Potluri, et al., 2020). This, therefore, necessitated the re-modeling of the propagation model, which now yielded the SEIR-HDVQ: Susceptible (S), exposed (E), Infected (I), Hospitalized (H), recovered (R), Death or death (D), Funeral (F), Vaccinated (V), and Quarantine (Q). Also, in designing the model, we considered that an insignificant number of recovered cases might still retain the virus in their body fluid, which has potency for infecting healthy cases (WHO, 2016), (UNCHC, 2017) (Thorson, et al., 2021). Since the interest of this study is to leverage the propagation model of the EVD for developing solving optimization algorithms, it became necessary to explore all factors supporting increased infection.

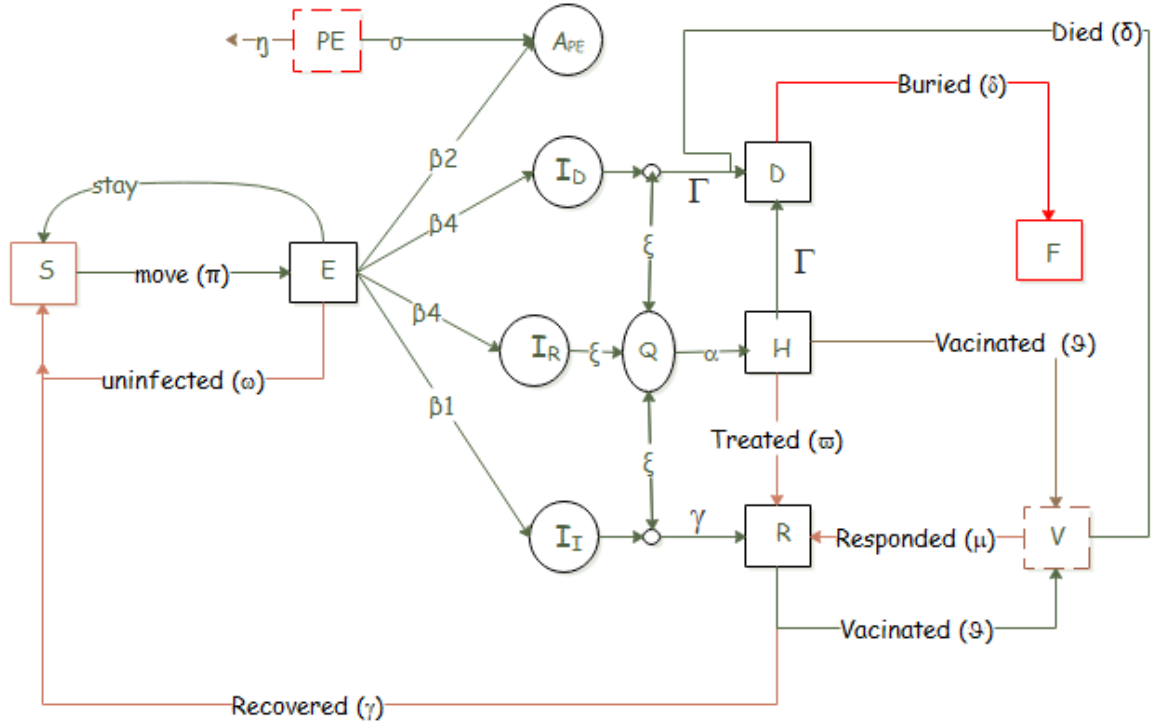

Figure 1: The SEIR-HDVQ propagation model of the proposed EOSA metaheuristic model

The model of the SEIR-HDVQ shown in Figure 1 and the listing of its parameters presented in Table 1. The propagation of EVD is assumed to provide a suitable manner for solving some optimization problems considering its aggressive mechanism overwhelming communities. The figure assumes a population of susceptible individuals whose exposure could trigger generating the population of other subgroups, though the reverse could also hold. Exposed individuals or contaminated environment or agent reservoirs can randomly draw arbitrary individuals from the susceptible into the categories of infected, which may be due to exposure to any individual from the subgroups of the infected. Subgroups of infected individuals are infected from the dead individual, infection from an infected individual who is alive, and infection from a recovered individual and infected from contaminated environment or agent-reservoir. We show that the virus has the potential of decaying in its contaminated environment. Furthermore, the propagation model shows that the infected cases could die without going to a hospital and may as well recover without hospitalization. An assumption made in this study is classifying every vaccinated case as hospitalized. Also, we assumed that both the hospitalized (H) and non-hospitalized cases could transit into the dead (D). At the same time, those recovered (R) from vaccination (V) are returned to the susceptible (S).

Table 1: Notations and description for variables and parameters for SEIR-HDVQ

| Symbols | Data Type | Descriptions |
|---------|-----------|--------------|
| S       | Vector    | Susceptible individuals |
| E       |           | Exposed individuals |

| I | | Infected individuals |
|---|---|---|
| H | | Hospitalized infected individuals |
| R | | Recovered infected individuals |
| D | | Diseased from infection individuals |
| V | | Vaccinated infected individuals |
| PE | | Agents capable of infecting individuals |
| $\pi$ | Scalar | Recruitment rate of susceptible human individuals |
| ŋ | | Decay rate of Ebola virus in the environment |
| $\alpha$ | | Rate of hospitalization of infected individuals |
| $\Gamma$ | | Disease-induced death rate of human individuals |
| $\beta_1$ | | Contact rate of infectious human individuals |
| $\beta_2$ | | Contact rate of pathogen individuals/environment |
| $\beta_3$ | | Contact rate of deceased human individuals |
| $\beta_4$ | | Contact rate of recovered human individuals |
| $\gamma$ | | Recovery rate of human individuals |
| $\tau$ | | Natural death rate of human individuals |
| $\delta$ | | Rate of burial of deceased human individuals |
| $\vartheta$ | | Rate of vaccination of individuals |
| $\varpi$ | | Rate of response to hospital treatment |
| $\mu$ | | Rate response to vaccination |
| $\xi$ | | Rate of quarantine of infected individuals |

The rates of change of variables or parameters applied in this study are summarized in Table 1. The values of most of these parameters are already predetermined by related studies on EVD, which this study simply leverages on their reported figures, as further discussed in Section 4.

### 3.2 Flowchart of EOSA

The SEIR-HDVQ propagation model presented in the last subsection motivated the design of the EOSA algorithm. This paper aims to adapt the propagation mechanism to the derivation and generation of potential search space for solving optimization problems using the proposed algorithm. The formalization of the EOSA algorithm is achieved from the following procedure:

1. Initialize all vector and scalar quantities which are individuals and parameters. Individuals in the sets: Susceptible (S), infected (I), recovered (R), dead (D), Vaccinated (V), Hospitalized (H), and Quarantine (Q) with their initial values.
2. Randomly generate the index case ($I_1$) from susceptible individuals.
3. Set the index case as the global best and current best, and compute the fitness value of the index case.
4. While the number of iteration is not exhausted and there exists at least an infected individual, then
   a. For each susceptible individual, generate and update their position based on their displacement. Note that the further an infected case is displaced, the more the number of infections, so that short displacement describes exploitation, otherwise exploration.
      i. Generate newly infected individuals (nI) base on (a).
      ii. Add the newly generated cases to I
   b. Compute the number of individuals to be added to H, D, R, B, V, and Q using their respective rates based on the size of I
   c. Update S and I base on nI.
   d. Select the current best from I and compare it with the global best.
   e. If the condition for termination is not satisfied, go back to step 6.
5. Return global best solution and all solutions.

In Figure 2 below, the flow chart of the proposed EOSA metaheuristic algorithm is shown.

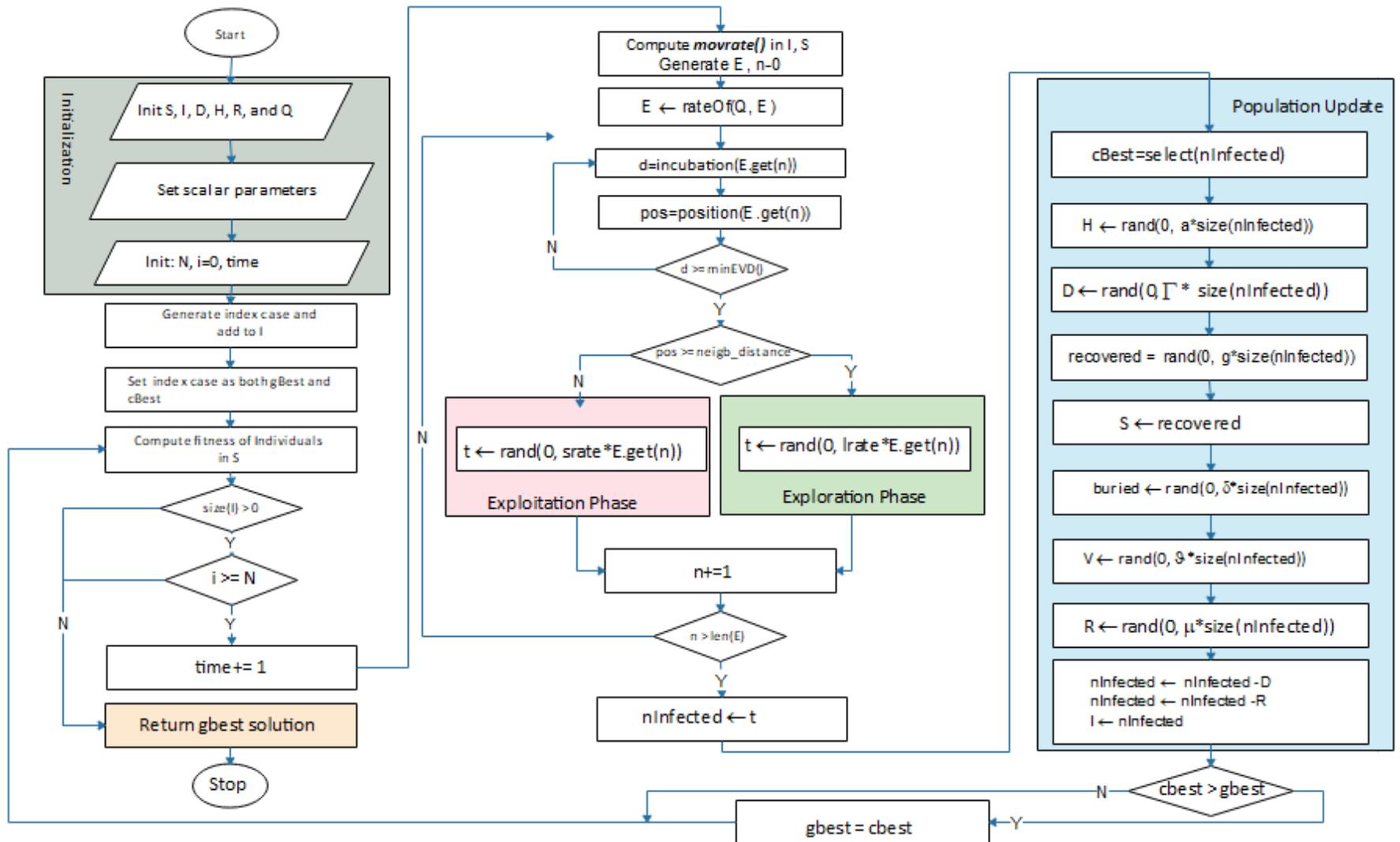

Figure 2: Flowchart of the proposed the EOSA metaheuristic algorithm

The flowchart presents a detailed flow of process and information as a build-up from the procedure described above. The detailing shows the various levels of initialization and conditional checking. Also, the computation leading to the exploration and exploitation stages of the proposed EOSA metaheuristic algorithm is detailed. Also, the procedure for update all subgroups are also identified. In the following subsection, the mathematical model applies to the flowchart of the algorithm presented and discussed.

### 3.3 Mathematical Model of EOSA

As earlier noted, we represent SEIR-HDVQ in this paper base on the definition of Susceptible (S), Infected (I), Hospitalized (H), Exposed (E), Recovered (R), Funeral (F), Dead (D), Vaccination (V), and Quarantine (Q). To update the positions of each exposed individual, Equation 1 applies:

$$mI_i^{t+1} = mI_i^t + \rho M(I) \tag{1}$$

where ρ represents the scale factor of displacement such individuals, $mI_i^{t+1}$ and $mI_i^t$ are respectively the updated and original position at time t, and t+1 is the current time. *M(I)* is the movement rate made by individuals, which is further defined thus:

$$M(I) = srate * rand\,(0,1) + M(Ind_{best}) \tag{2}$$

$$M(s) = lrate * rand\,(0,1) + M(Ind_{best}) \tag{3}$$

The exploitation stage of the EOSA optimization algorithm assumes that the infected individual either stays within a distance of zero (0), or is displaced within a limit not exceeding *srate* denoting short distance movement. The exploration phase of the algorithm assumes that the infected individual has moved beyond the normal neighborhood range *lrate*. The consideration in this study is that the farther the displacement, the more the number of contacts exposed to the infection. Equations 2 and 3, therefore, ensure that the movement of each individual in consideration is appropriately assigned. Both *srate* and *lrate* are regulated by *neighborhood* parameter. If *neighborhood* is over 0.5, we assume the individual has moved beyond the *neighborhood* otherwise, remains within the *neighborhood*.

**Initialization of Susceptible population**: At the beginning, an initial population is generated by means of random number distribution whose initial positions are all zero (0), so that *i*[th] individual is generated as shown in Equation 4. The function *rand (0, 1)* generates uniformly distributed values, the variable *i*, and *$U_i$* and *$L_i$* denote the upper and lower bounds respectively for the *i*[th] individual, ranges from 1,2,3…. N, where is the population size.

$$individual_i = L_i + rand(0,1) * (U_i + L_i) \tag{4}$$

The selection of the current best is carried out on the set of infected individuals in time t. meanwhile, the selection of the global best is based on the following:

$$bestS = \begin{cases} gBest, & fitness(cBest) < fitness(gBest) \\ cBest, & fitness(cBest) \geq fitness(gBest) \end{cases} \tag{5}$$

where **bestS**, **gBest** and **cBest** all denotes the best solution, global best solution, and current best solution at time *t*; *fitness(.)* represents the objective function applied to the problem. We distinguish **gBest** and **cBest** as infected individuals who **Superspreader** and **Spreader** of the Ebola virus, respectively.

Update of Susceptible (S), Infected (I), Hospitalized (H), Exposed (E), Vaccinated (V), Recovered (R), Funeral (F), Quarantine (Q), and dead (D) uses a system of ordinary differential equations based on those in (Berge, Lubuma, Moremedi, Morris, & Kondera-Shava, 2017) (Tanade, Pate, Paljug, Hoffman, & Wang, 2019). Differential calculus is a branch of calculus that is a branch in mathematics, where the former deals with the rate of change of one quantity concerning another, while the latter deals with finding different properties of integrals and derivatives. The application of differential calculus, in our case, intends to obtain the rates of change of quantities S, I, H, R, V D, and Q with respect to time *t*. Hence, the Equations 6, 7, 8, 9, 10, 11, 12 for S, I, H, R, V D, and Q respectively as follows:

$$\frac{\partial S(t)}{\partial t} = \pi - (\beta_1 I + \beta_3 D + \beta_4 R + \beta_2 (PE)\eta)S - (\tau S + \Gamma I) \tag{6}$$

$$\frac{\partial I(t)}{\partial t} = (\beta_1 I + \beta_3 D + \beta_4 R + \beta_2 (PE)\lambda)S - (\Gamma + \gamma)I - (\tau)S \tag{7}$$

$$\frac{\partial H(t)}{\partial t} = \alpha I - (\gamma + \varpi)H \tag{8}$$

$$\frac{\partial R(t)}{\partial t} = \gamma I - \Gamma R \tag{9}$$

$$\frac{\partial V(t)}{\partial t} = \gamma I - (\mu + \vartheta)V \tag{10}$$

$$\frac{\partial D(t)}{\partial t} = (\tau S + \Gamma I) - \delta D \tag{11}$$

$$\frac{\partial Q(t)}{\partial t} = (\pi I - (\gamma R + \Gamma D)) - \xi Q \tag{12}$$

We shall first assume that each of Equations (6 -11) is a *scalar function*, meaning that it has one number as a value, which can be represented as a real value. This is not far removed from some common scalar differential equations and their corresponding *f* functions, which include exponential growth of money or populations governed by scalar differential equations: *u'=αu*, where *u is the* growth rate.

We determine the rate of change of the population of susceptible individuals and then apply it to the current size of the susceptible vector to obtain the number of susceptible individuals at time *t*. The same procedure is applied to compute the set of individuals in vectors I, H, R, V, D, and Q using rates described in Table 1. This study assumes the initial conditions S(0) = S0, I(0) = I0, R(0) = R0, D(0) = D0, P(0) = P0, and Q(0) = Q0 where our *t* follows after the epoch, and δ in Equation (11) is for the burial rate. Equation (12) models the rate of quarantine of infected cases of Ebola.

### 3.4 Algorithm Design of EOSA
The pseudo-code of the proposed EOSA metaheuristic algorithm is shown in Algorithm 1. Lines 1-7 of the algorithm show the initialization phase of parameters used in subsequent lines.

**Algorithm 1:** Algorithm of the EOSA metaheuristic algorithm

**Result:** Best solution
**Input:** objfunc, lb, ub, epoch, psize, evdincub
**Output:** solution, gbest

1 $S, E, I, H, R, V, Q, sols \leftarrow \emptyset$;
2 Initialize S finite set $S = \{ind_1, ind_2, \ldots, ind_n\}$ ;
3 $S \leftarrow createSusceptibleIndvd(psize, S), Eq.4$;
4 $time \leftarrow 0$;
5 $icase \leftarrow generatedIndexCase()$;
6 $gbest, cbest \leftarrow icase$;
7 **while** $e \leq epoch \land len(I) > 0$ **do**
8    $Q \leftarrow rand(0, Eq.12 \times I)$;
9    $fracI = I - Q$;
10    **for** $i \leftarrow 1$ **to** $len(fracI)$ **do**
11       $pos_i \leftarrow movrate()$   $using\ Eq.1$;
12       $d_i \leftarrow rand()$;
13       **if** $d_i > evdincub$ **then**
14          $neighborhood \leftarrow prob(pos_i)$;
15          **if** $neighborhood < 0.5$ **then**
16             $tmp \leftarrow rand(0, Eq.7 \times I \times srate)$;
17          **end**
18          **else**
19             $tmp \leftarrow rand(0, Eq.7 \times I \times lrate)$;
20          **end**
21          $newI+ \leftarrow tmp$;
22       **end**
23       $I+ \leftarrow newI$;
24    **end**
25    $h \leftarrow rand(0, Eq.8 \times I), H+ \leftarrow h$;
26    $r \leftarrow rand(0, Eq.9 \times I), R+ \leftarrow r$;
27    $v \leftarrow rand(0, Eq.10 \times h), V+ \leftarrow v$;
28    $d \leftarrow rand(0, Eq.11 \times I), D+ \leftarrow d$;
29    $I+ \leftarrow I - add(r, d)$;
30    $S+ \leftarrow r$;
31    $S- \leftarrow d$;
32    $cbest = fitness(objfunc, I)$;
33    **if** $cbest > gbest$ **then**
34       $gbest = cbest$;
35       $sols \leftarrow gbest$;
36    **end**
37 **end**
38 **return** gbest, sols;

To demonstrate that not all infected cases have the potency for recruiting new infected individuals, Line 8 shows that a sample is drawn into quarantine so that the remaining fraction of I infect the S population. On Lines 10-24, new infections are generated from S and then added to I. Since R, V, H, and V are only derivable from I, Lines 25-29 of Algorithm 1 generate their individuals using the appropriate equations. Logically, recovered and dead cases need to be removed from I before the next period. Also, recovered cases are added back to S while dead individuals are replaced in S with new cases. These are all model in Lines 29-31. Finally, the best solution is computed, the termination criterion is checked so that when it is satisfied, the algorithm terminates otherwise return to Line 7.

The overview of the proposed metaheuristic algorithm, EOSA, has been defined and designed in this Section. To demonstrate its practicability, we follow in the next Section for experimental setup, configurations, and parameter definition. These parameters provide details on those used in Algorithm 1.

## 4. Parameter Settings and Experimental Setup

This section is focused on reporting the environment for carrying out very detailed experimentation of the EOSA algorithm. First, we show control parameter settings and variable assignment, then list the benchmark functions applied to the algorithm, and then finally detail the evaluation criteria.

### 4.1 Configuration of experimental setup

Exhaustive experimentation for evaluating the proposed EOSA described in Algorithm 1 was carried out in a workstation environment with the following configurations: Intel (R) Core i5-7500 CPU 3.40GHz, 3.41GHz; RAM of 16 GB; 64-bit Windows 10 OS for each configuration of the system on the network. A total of ten (10) existing metaheuristic algorithms were compared with the proposed EOSA algorithm. To ensure that there is fairness in the execution of each algorithm, this study executed each algorithm twenty (20) times. Also, a total of five hundred (500) epochs were covered in each run of 20 for each algorithm. The runs of 20 for each algorithm allow computing the average values for all metrics in each 500 epochs.

### 4.2 Parameters of EOSA metaheuristic algorithm

The design and selection of EOSA's parameters and their corresponding values assumed the natural definitions generated from those reported by WHO. This study adopted the rates reported in studies that have carried out an extensive evaluation of the SEIR models they proposed. These studies relied on the WHO data for the evaluation of their models. All these parameters have been described and applied in Sections 3.1, 3.2, and 3.3, where the SEIR-HDVQ model was discussed.

Table 2: Notations and description for variables and parameters for SEIR-HDVQ

| Symbols | Descriptions | Range |
| --- | --- | --- |
| $\pi$ | Recruitment rate of susceptible human individuals | Variable |
| $\eta$ | Decay rate of Ebola virus in the environment | $(0, \infty)$ |
| $\alpha$ | Rate of hospitalization of infected individuals | $(0, 1)$ |
| $\Gamma$ | Disease-induced death rate of human individuals | $[0.4, 0.9]$ |
| $\beta_1$ | Contact rate of infectious human individuals | Variable |
| $\beta_2$ | Contact rate of pathogen individuals/environment | Variable |
| $\beta_3$ | Contact rate of deceased human individuals | Variable |
| $\beta_4$ | Contact rate of recovered human individuals | Variable |
| $\gamma$ | Recovery rate of human individuals | $(0, 1)$ |
| $\tau$ | Natural death rate of human individuals | $(0, 1)$ |
| $\delta$ | Rate of burial of deceased human individuals | $(0, 1)$ |
| $\vartheta$ | Rate of vaccination of individuals | $(0, 1)$ |
| $\varpi$ | Rate of response to hospital treatment | $(0, 1)$ |
| $\mu$ | Rate response to vaccination | $(0, 1)$ |
| $\xi$ | Rate of quarantine of infected individuals | $(0, 1)$ |

In Table 2, the initial value for each parameter is defined. Considering the stochastic nature of EOSA, which is characteristic of biology-based optimization algorithms, some of the parameters assumed random values within the range of zero's (0) and 1's (1).

**4.3 Benchmark functions**

As an effort to evaluate the effectiveness of the proposed EOSA metaheuristic algorithm, this study curated forty-seven (47) standard and high dimensional functions for this purpose. These functions are listed in Table 3 and are subsequently used for observing the performance of EOSA with the hope of comparing its performance with those of a similar bioinspired metaheuristic algorithm. To give a proper presentation of the functions, we curated the names with the mathematical model representation of the functions alongside their range. These classical functions were also combined with the CEC'2010 and CEC'2017 benchmark functions to achieve very exhaustive experimentation.

Table 3: Standard benchmark functions used for the experimentation: Dimensions (D), Multimodal (M), non-separable (N), Unimodal (U), Separable (S)

| ID | Function name | Range | Model of the function | D | Type | Min |
|---|---|---|---|---|---|---|
| F1 | Ackley | $[-32, 32]$ | $f(x) = -20e^{\left(-0.2\sqrt{\frac{1}{n}\sum_1^n x_i^2}\right)} - -e^{\left(\frac{1}{n}\sum_1^n \cos(2\pi x_i)\right)} + 20 + e^{(1)}$ | 30 | MN | 0 |
| F2 | Alpine | $[-10, 10]$ | $f(x) = \sum_{i=1}^{n} \lvert x_i \sin(x_i) + 0.1 x_i \rvert$ | N | MN | 0 |
| F3 | Brown | $[-1, 4]$ | $f(x) = \sum_{i=1}^{n-1} (x_i^2)^{(x_{i+1}^2+1)} + (x_{i+1}^2)^{(x_i^2+1)}$ | N | UN | 0 |
| F4 | Bent Cigar | $[-100,100]$ | $f_{20}(x) = x_1^2 + 10^6 \sum_{i=2}^{D} x_i^2$ | N | MS | 0 |
| F5 | Composition1 | $[-100,100]$ | g1=Rosenbrock's Function F29<br>g2=High Conditioned Elliptic Function F15<br>g3=Rastrigin's Function F27 | 5 | | |
| F6 | Composition2 | $[-100,100]$ | g1=Ackley's Function F1<br>g2=High Conditioned Elliptic Function F15<br>g3=Griewank Function F10<br>g4=Rastrigin's Function F27 | 3 | | |
| F7 | Dixon and Price | $[-10, 10]$ | $f_{18}(x) = 10^6 x_1^2 \sum_{i=2}^{D} x_i^2$ | 30 | UN | 0 |
| F8 | Discus Function | $[-100, 100]$ | $f(x) = (x_1 - 1)^2 + \sum_{i=2}^{n} i(2x_i^2 - x_{i-1})^2$ | N | U | |
| F9 | Fletcher–Powel | $[-100, 100]$ | $f(x) = 100\left\{[x_3 - 10\theta(x_1,x_2)]^2 + \left(\sqrt{x_1^2 + x_2^2} - 1\right)^2\right\} + x_3^2$<br>Where $2\pi\theta(x_1,x_2) = \begin{cases} \tan^{-1}\frac{x_2}{x_1}, & if\, x_1 \geq 0 \\ \pi - \tan^{-1}\frac{x_2}{x_1}, & otherwise \end{cases}$ | N | MN | 0.0001 |
| F10 | Griewank | $[-600, 600]$ | $f(x) = 1 + \sum_{i=1}^{n} \frac{x_i^2}{1400} - \prod_{i=1}^{n} \cos(\frac{x_i}{\sqrt{i}})$ | 30 | MN | 0 |
| F11 | Generalized Penalized Function 1 | $[-50, 50]$ | $f(x) = \frac{\pi}{n} X \left\{10\sin^2(\pi y_i) + \sum_{i=1}^{n-1}(y_i - 1)^2[1 + 10\sin^2(\pi y_{i+1})] + (y_n - 1)^2\right\}$<br>$+ \sum_{i=1}^{n} u(x_i, a, k, m)$ | n | M | 0 |
| | | | Where $y_i = 1 + \frac{1}{4}(x_i + 1)$, $u(x_i, a, k, m) = \begin{cases} k(x_i - a)^m & \text{if } x_i > a \\ 0 & \text{if } -a \leq x_i \leq a \\ k(-x_i - a)^m & \text{if } x_i < -a \end{cases}$<br>a=10, k=100, m=4 | | | |
| F12 | Generalized Penalized Function 2 | [-5.12, 5.12] | $f(x) = 0.1 \, X \left\{ \sin^2(3\pi x_1) + \sum_{i=1}^{n-1}(x_i - 1)^2 [1 + \sin^2(3\pi x_{i+1})] + (x_n - 1)^2[1 + \sin^2(2\pi x_n)] \right\}$<br>$+ \sum_{i=1}^{n} u(x_i, a, k, m)$<br>Where $u(x_i, a, k, m) = \begin{cases} k(x_i - a)^m & \text{if } x_i > a \\ 0 & \text{if } -a \leq x_i \leq a \\ k(-x_i - a)^m & \text{if } x_i < -a \end{cases}$<br>a=5, k=100, m=4 | N | M | 0 |
| F13 | Holzman 2 function | [-100,100] | $f(x) = \sum_{i=1}^{n} i x_i^4$ | N | | |
| F14 | HGBat | [-100,100] | $f_{23}(x) = \left\| \left(\sum_{i=1}^{D} x_i^2\right)^2 - \left(\sum_{i=1}^{D} x_i\right)^2 \right\|^{1/2} + \left(0.5 \sum_{i=1}^{D} x_i^2 + \sum_{i=1}^{D} x_i\right)/D + 0.5$ | 30 | M | |
| F15 | High Conditioned Elliptic | [-100,100] | $f_{23}(x) = \sum_{i=1}^{D} (10^6)^{\frac{i-1}{D-1}} x_i^2$ | N | | |
| F16 | Hybrid1 | [-100,100] | g1: Zakharov Function F45<br>g2: Rosenbrock Function F29<br>g3: Rastrigin's Function F27 | 3 | UN | 0 |
| F17 | Hybrid2 | [-100,100] | g1: High Conditioned Elliptic Function F15<br>g2: Ackley's Function F1<br>g3: Rastrigin's Function F27<br>g4: HGBat Function F14<br>g4: Discus Function F8 | 3 | MN | 0 |
| F18 | Inverted Cosine Mixture | [-1,1] | $f_{14}(x) = 0.1n - (0.1 \sum_{i=1}^{n} \cos(5\pi x_i) - \sum_{i=1}^{n} x_i^2)$ | N | MS | -0.1x(n) |
| F19 | Lévy 3 function | $[-10, 10]$ | $f(x) = \sum_{i=1}^{n-1}\left[0.5 + \frac{\sin^2\left(\sqrt{100x_i^2 + x_{i+1}^2}\right) - 0.5}{1 + 0.001(x_i^2 - 2x_ix_{i+1} + x_{i+1}^2)^2}\right]$ | N | | |
| F20 | Levy | $[-10, 10]$ | $f_{12}(x) = \sum_{i=1}^{n}(x_i - 1)^2\left[\sin^2(3\pi x_{i+1})\right] + \sin^2(3\pi x_1) + |x_n - 1|[1 + \sin^2(3\pi x_n)]$ | 2 | MN | 0 |
| F21 | Levy and Montalo | $[-5, 5]$ | $f_{17}(x) = 0.1\left(\sin^2(3\pi x_1)\right)$ $+ \sum_{i=1}^{n}(x_i - 1)^2\left(1 + \sin^2(3\pi x_{1+1})\right) + (x_n - 1)^2\left(1 + \sin^2(2\pi x_n)\right)$ | N | MS | 0 |
| F22 | Noise | $[-1.28, 1.28]$ | $f_7(x) = \sum_{i=1}^{n} x_i^4 + random[0,1)$ | N | | |
| F23 | Pathological function | $[-100, 100]$ | $f(x) = \sum_{i-1}^{5} i\cos((i-1)x_1 + i)\sum_{j=1}^{5} j\cos\left((j+1)x_1 + j\right)$ | N | MN | 0 |
| F24 | Perm | $[-20, 20]$ | $f(x) = \sum_{k=1}^{n}\left[\sum_{i=1}^{n}(i_k + \beta)\left(\left(\frac{x_i}{i}\right)^k - 1\right)\right]^2$ | N | MN | 0 |
| F25 | Powel | $[-4, 5]$ | $f(x) = (x_1 + 10x_2)^2 + 5(x_3 + x_4)^2 + (x_2 - 2x_3)^4 + 10(x_1 - x_4)^4$ | N | UN | 0 |
| F26 | Quartic | $[-128, 128]$ | $f_6(x) = \sum_{i=1}^{n} ix_i^4$ | 30 | MS | 0 |
| F27 | Rastrigin | $[-5.12, 5.12]$ | $f_9(x) = \sum_{i=1}^{n}[x_i^2 - 10\cos(2\pi x_i) + 10]$ | 30 | MN | 0 |
| F28 | Rotated hyperellipsoid | $[-100, 100]$ | $f_3(x) = \sum_{i=1}^{n}\left(\sum_{j=1}^{i} x_j\right)$ | N | U | 0 |
| F29 | Rosenbrock | $[-30, 30]$ | $f(x) = \sum_{i=1}^{n-1}[100\,(x_{i+1} - x_i^2)^2 + (x_i - 1)^2]$ | 30 | UN | 0 |
| F30 | Schwefel 2.26 | $[-500, 500]$ | $f(x) = \sum_{i=1}^{n}[-x_i \sin(\sqrt{|x_i|})]$ | N | MS | −418.983 |
| F31 | Schwefel 1.2 | $[-100, 100]$ | $f(x) = \sum_{i=1}^{n}\left(\sum_{j=1}^{i} x_j\right)^2$ | 30 | UN | 0 |
| F32 | Schwefel 2.22 | $[-100, 100]$ | $$f(x) = \sum_{i=1}^{n}|x_i| + \prod_{i=1}^{n}|x_i|$$ | 30 | UN | 0 |
| F33 | Schwefel 2.21 | $[-100, 100]$ | $f(x) = \max\{|x_i|, 1 \leq i \leq n\}$ | N | US | 0 |
| F34 | Sphere | $[-100, 100]$ | $$f_1(x) = \sum_{i=1}^{n} x_i^2$$ | 30 | US | 0 |
| F35 | Step | $[-100, 100]$ | $$f(x) = \sum_{i=1}^{n}(floor(x_i) + 0.5)^2$$ | 30 | US | 0 |
| F36 | Sum/SumSquares Function | $[-10, 10]$ | $$f(x) = \sum_{i=1}^{n} i x_i^2$$ | 30 | US | 0 |
| F37 | Sum-Power | $[-1, 1]$ | $$f_8(x) = \sum_{i=1}^{n}|x_i|^2$$ | N | US | 0 |
| F38 | Sum of Different Power | $[-100,100]$ | $$f_{21}(x) = \sum_{i=1}^{d}|x_i|^{i+1}$$ | N | US | 0 |
| F39 | SR-F4 | $[-100,100]$ | Shifted and Rotated Bent Cigar Function | N | | |
| F40 | SR-F38 | $[-100,100]$ | Shifted and Rotated Sum of Different Power Function | N | | |
| F41 | SR-F45 | $[-100,100]$ | Shifted and Rotated Zakharov Function | N | | |
| F42 | SR-F29 | $[-100,100]$ | Shifted and Rotated Rosenbrock's Function | N | MN | 0 |
| F43 | SR-F27 | $[-100,100]$ | Shifted and Rotated Rastrigin's Function | N | MS | 0 |
| F44 | Wavy 1 | $[-100,100]$ | $$f(x) = \sum_{i=1}^{n} x_i^2 + (\sum_{i=1}^{n} 0.5ix_i)^2 + (\sum_{i=1}^{n} 0.5ix_i)^4$$ | 2 | MS | 0 |
| F45 | Zakharov | $[-5, 10]$ | $$f(x) = \frac{1}{n}\sum_{i=1}^{n} 1 - \cos(10x_i)\, e^{-\frac{1}{2}x_i^2}$$ | 10 | UN | 0 |
| F46 | Salomon | $[-100, 100]$ | $$f_{19}(x) = 1 - \cos\left(2\pi\sqrt{\sum_{i=1}^{n} x_i^2}\right) + 0.1\sqrt{\sum_{i=1}^{n} x_i^2}$$ | N | MN | 0 |
| F47 | Weierstrass Function | $[-0.5, 0.5]$ | $$f(x) = \sum_{i=1}^{D}(\sum_{i=0}^{20}[0.5^k \cos(2\pi \cdot 3^k(x_i + 0.5))])$$ | 50 | MN | 0 |

The list of common standard literature benchmark functions outlined in Table 3 is used to test and evaluate the quality of the proposed EOSA algorithm. The literature provides a classification for each of these functions to demonstrate the expected performance they are to show. Whereas many test functions are continuous, they are further categorized into four (4). In the first class are test functions characterized by unimodal, convex, and multidimensional forms. They represent a class of test functions with interesting functions with cases capable of slowing down convergence or even yielding a poor convergence. The resulting convergence trails from such a slow pace to a single global extremum. The second class consists of test functions of type multimodal, two-dimensional with a small number of local extremes. This category of test functions appeals to situations where we intend to test the quality of standard optimization procedures in an anticipated hostile environment. This hostile environment describes problem domains with only a few local extremes with a single global one.

The third and fourth classes collect a list of test functions known as multimodal, two-dimensional with a huge number of local extremes, and multimodal, multidimensional, with a huge number of local extremes, respectively. It has been shown that these test functions work well for situations where the quality of intelligent and resistant optimization algorithms are tested (Jamil & Yang, 2013), (Yang, 2010), (Molga & Smutnicki, 2020), (Digalakis & Margariti, 2014), (Hussain, Salleh, Cheng, & Naseem, 2017). The proposed EOSA method was also tested on 30 benchmark functions from IEEE CEC.

### 4.4 Evaluation Method
The following criteria are considered in the performance evaluation: mean, median, standard deviation, maximum values, minimum or worst values, average values, overall convergence time, and average execution time. In addition to these, we applied the outcome of the proposed EOSA and related optimization algorithms to statistical tests to evaluate their performance in terms of convergence to determine algorithms capable of generating similar final solutions. These metrics provide an unbiased platform for the comparison of algorithms discussed.

## 5. Result and Discussion
A detailed performance evaluation of the experiment's outcome carried out in Section 4 is presented and discussed in this Section. First, we study the performance of the proposed SEIR-HDVQ model to determine how effectively the model was able to describe the natural phenomenon associated with it. Thereafter, the performance of EOSA was compared with all selected metaheuristic algorithms experimented with. Performance evaluation is done using the values obtained while applying the optimization algorithms to the test functions discussed earlier. Finally, a discussion on the findings and suitability of the proposed EOSA algorithm is presented.

### 5.1 Simulation of EVD Propagation Based on SEIR-HDVQ model

This subsection presents the result of simulating the proposed SEIR-HDVQ model with the same randomly generated data applied to the experimentation of EOSA. This result is reported to investigate the rate at which the SEIR-HDVQ model represents corresponding curves obtained in the real-life propagation model for the EVD and EBOV.

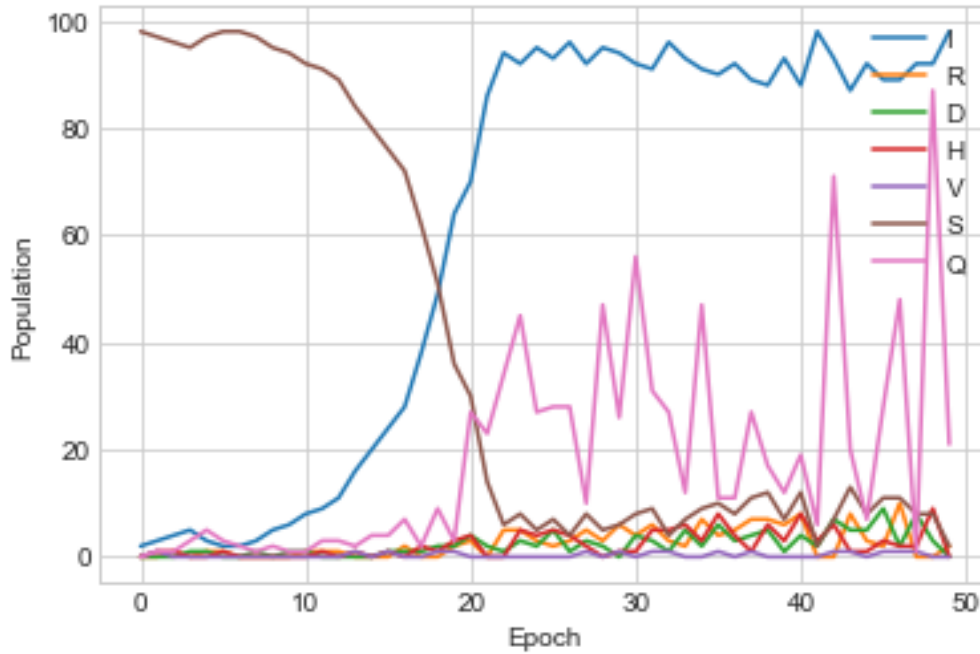

**Figure 3:** An estimated Ebola virus and disease propagation curve based on the simulation with randomly generated data while experimenting with the EOSA optimization algorithm. The curve illustrates variations in the values of Susceptible (S), Infected (I), Recovered (R), Hospitalized (H), Dead (D), Vaccinated (V), and Quarantine (Q) using the SEIR-HDVQ model

In Figure 3, the curves for Susceptible (S), Infected (I), Recovered (R), Hospitalized (H), Dead (D), Vaccinated (V), and Quarantine (Q) are captured so that they show the rate at which each compartment rises and fall within a period of fifty (50) epochs. In the figure, we observed that in the early phase of the outbreak, the infection rate rose against the susceptible population. The figure also revealed the response to the rising infection through quarantine measure, such that as infection rose, the number of quarantine individuals also increased – a measure to stem the outbreak. Meanwhile, we noticed that recovery and death rate curves wobbled along the curve of infection and quarantine.

### 5.2 Performance of EOSA with Similar Metaheuristic Algorithms Using Classical Benchmark functions

The performance of EOSA was compared with seven (7) different optimization algorithms, namely Artificial Bee Colony (ABC), Whale Optimization Algorithm (WOA), Butterfly Optimization Algorithm (BOA), Particle Swarm Optimization (PSO), Differential Evolution (DE), *Genetic Algorithm* (GA), and Henry gas solubility optimization algorithm (HGSO). The experimentation, which was executed for five hundred (500) iterations and twenty (20) different runs, applied forty-seven (47) standard benchmark functions.

In Table 4, the outcome for the best, worst, mean, median, and standard deviation for each of the 47 functions are listed. An overview of the result showed that although EOSA outperformed most of the algorithms in most cases for the 47 functions, some interesting differences were noticed to group the outcome into two. Whereas EOSA demonstrated a very close performance compared with ABC, WOA, BOA, and PSO, we observed that EOSA's outcome compared with DE, GA, and HGSO was significantly better. For example, for F1-4, F6-7, F12, F14-15, F18, F20-23, F2, F29-30, F32-36, F38, F40-43, and F46-47, EOSA clearly achieved the best values in all cases. However, in F5, F9-11, F13, F16-17, F19, and F25-26, EOSA was outperformed based on WOA and BOA, ABC, WOA, and PSO ABC, BOA and PSO, ABC and PSO, ABC, ABC, WOA, and BOA, respectively. Meanwhile, we found some situations for the 47 functions where there was no clear superiority of EOSA over similar algorithms, neither were the similar algorithms able to demonstrate clear superiority. These cases are found in F28, F31, F37, F39, F44, and F45, where we observed that EOSA was beaten by ABC and PSO, beaten by WOA, BOA and PSO, ABC, matching values in ABC and PSO, beaten by WOA, and beaten by ABC, WOA, and PSO respectively

The values obtained for the worst, as shown in Table 4 for the 47 functions, revealed that there is an intense competition between EOSA and ABC, WOA, BOA, and PSO. We discovered that only in the cases of F3, F6, F12, F18, F22-23, F27, F29, F28, and F40-41was the values of EOSA better than those earlier listed and was also completely outperformed those same algorithms in the cases of F13, F21, F25, and F32. We, however, noticed that the discrepancies reported by these algorithms in comparison with EOSA were not significantly large. Meanwhile, DE, GA, and HGSO have maintained a significant variation in the values obtained for those of best and worst computations. This implied that the proposed EOSA algorithm and its competitive related algorithms (ABC, WOA, BOA, and PSO) all significantly performed well over DE, GA, and HGSO.

**Table 4:** Comparison of best, worst, mean, median, and standard deviation (stdev) values for ABC, WOA, BOA PSO, DE, GA, and HGSO metaheuristic algorithms using the classical benchmark functions over 500 runs and 100 population size

| Function | Metrics | Compared Algorithms | | | | | | | |
|---|---|---|---|---|---|---|---|---|---|
| | | ABC | WOA | BOA | PSO | EOSA | DE | GA | HGSO |
| F1 | best | 0.046635003 | 0.046548228 | 0.046606998 | 0.046595289 | 0.046579079 | 19.96184399 | 9.834179892 | 4.44E-16 |
| | worst | 20.90485862 | 0.046548228 | 0.046606998 | 0.046595289 | 0.046590768 | 20.91052673 | 19.84903108 | 17.47294741 |
| | mean | 19.35207569 | 0.046548228 | 0.046606998 | 0.046595289 | 0.046579273 | 20.0105497 | 10.33766475 | 0.224608623 |
| | median | 19.21059815 | 0.046548228 | 0.046606998 | 0.046595289 | 0.046579079 | 19.96347193 | 10.09803718 | 4.44E-16 |
| | stdev | 0.943908994 | 4.86E-18 | 5.20E-18 | 5.20E-18 | 1.24E-06 | 0.146978912 | 0.961561963 | 1.441674873 |
| F2 | best | 0.002749312 | 0.002795749 | 0.002762226 | 0.002778424 | 0.002735207 | 202.2092691 | 39.09706932 | 1.15E-59 |
| | worst | 243.8707278 | 0.002795749 | 0.002762226 | 0.002778424 | 0.002786444 | 243.8368578 | 181.9837599 | 151.3617491 |
| | mean | 33.43141849 | 0.002795749 | 0.002762226 | 0.002778424 | 0.002735311 | 223.0877822 | 43.99604306 | 1.067724803 |
| | median | 7.44293501 | 0.002795749 | 0.002762226 | 0.002778424 | 0.002735207 | 219.8379837 | 41.29894943 | 1.70E-31 |
| | stdev | 52.1262757 | 4.12E-19 | 3.90E-19 | 4.34E-19 | 2.29E-06 | 13.7309761 | 10.72063849 | 10.27440175 |
| F3 | best | 0.000414246 | 0.000406423 | 0.00041358 | 0.00040821 | 0.000341339 | 506.4395978 | 923.2199937 | 200 |
| | worst | 1459.126847 | 0.000406423 | 0.00041358 | 0.00040821 | 0.000402738 | 1511.839293 | 1242.053574 | 937.124982 |
| | mean | 290.8930668 | 0.000406423 | 0.00041358 | 0.00040821 | 0.000341484 | 766.5204487 | 940.9341817 | 203.9456023 |
| | median | 202.646981 | 0.000406423 | 0.00041358 | 0.00040821 | 0.000341339 | 689.7884683 | 931.4994158 | 200 |
| | stdev | 219.6111673 | 3.79E-20 | 6.23E-20 | 7.32E-20 | 2.87E-06 | 242.3395941 | 28.42301313 | 44.96951031 |
| F4 | best | 2.47E-12 | 2.48E-12 | 2.50E-12 | 2.44E-12 | 2.44E-12 | 1.46E+11 | 4106464761 | 8.23E-104 |
| | worst | 2.60168E+11 | 2.48E-12 | 2.50E-12 | 2.44E-12 | 2.44E-12 | 2.58E+11 | 1.35E+11 | 1.07E+11 |
| | mean | 2.04696E+11 | 2.48E-12 | 2.50E-12 | 2.44E-12 | 2.44E-12 | 2.06E+11 | 5647478875 | 489880801.7 |
| | median | 2.00955E+11 | 2.48E-12 | 2.50E-12 | 2.44E-12 | 2.44E-12 | 2.09E+11 | 4340687923 | 1.73E-49 |
| | stdev | 13029284872 | 4.24E-28 | 3.23E-28 | 3.43E-28 | 4.27E-17 | 39224298697 | 7415345765 | 5813315095 |
| F5 | best | 1.26E-06 | 1.23E-06 | 1.23E-06 | 1.24E-06 | 1.24E-06 | 293464.1073 | 10387.08092 | 0 |
| | worst | 513842.7078 | 1.23E-06 | 1.23E-06 | 1.24E-06 | 1.24E-06 | 510854.4233 | 271266.3555 | 198959.4441 |
| | mean | 408834.0144 | 1.23E-06 | 1.23E-06 | 1.24E-06 | 1.24E-06 | 389379.6679 | 13319.70525 | 1007.87092 |
| | median | 401536.9082 | 1.23E-06 | 1.23E-06 | 1.24E-06 | 1.24E-06 | 370976.4288 | 10947.76684 | 0 |
| | stdev | 25604.25205 | 1.16E-22 | 1.59E-22 | 1.16E-22 | 9.72E-11 | 70462.61054 | 14783.10391 | 11466.50545 |
| F6 | best | 2.40E-18 | 2.44E-18 | 2.37E-18 | 2.44E-18 | 2.37E-18 | 1.49E+17 | 53323511.38 | 44.10261447 |

|  |  | Compared Algorithms | | | | | | | |
| --- | --- | --- | --- | --- | --- | --- | --- | --- | --- |
| Function | Metrics | ABC | WOA | BOA | PSO | EOSA | DE | GA | HGSO |
|  | worst | 1.45E+17 | 2.44E-18 | 2.37E-18 | 2.44E-18 | 2.40E-18 | 1.49E+17 | 2.52E+16 | 1.04E+16 |
|  | mean | 6.40E+16 | 2.44E-18 | 2.37E-18 | 2.44E-18 | 2.37E-18 | 1.49E+17 | 5.83E+13 | 3.01E+13 |
|  | median | 5.70E+16 | 2.44E-18 | 2.37E-18 | 2.44E-18 | 2.37E-18 | 1.49E+17 | 1670980321 | 44.12595826 |
|  | stdev | 1.59E+16 | 2.31E-34 | 2.50E-34 | 3.85E-34 | 1.55E-21 | 19.2 | 1.14E+15 | 5.15E+14 |
| F7 | best | 2.86E-12 | 2.82E-12 | 2.76E-12 | 2.86E-12 | 2.57E-12 | 2.71E-146 | 865.7595089 | 1.02E-241 |
|  | worst | 100289061.5 | 2.82E-12 | 2.76E-12 | 2.86E-12 | 2.80E-12 | 24647958.47 | 652837.5949 | 54779492.97 |
|  | mean | 256702.4063 | 2.82E-12 | 2.76E-12 | 2.86E-12 | 2.57E-12 | 85506.69928 | 26635.78996 | 114542.9249 |
|  | median | 548.2299318 | 2.82E-12 | 2.76E-12 | 2.86E-12 | 2.57E-12 | 3.40E-69 | 5236.830702 | 1.47E-122 |
|  | stdev | 4643773.174 | 3.64E-28 | 2.83E-28 | 3.43E-28 | 1.06E-14 | 1297580.095 | 84911.8986 | 2464983.02 |
| F8 | best | 1.01E-10 | 1.02E-10 | 1.02E-10 | 1.01E-10 | 1.01E-10 | 152250.3059 | 11996.3679 | 1.22E-109 |
|  | worst | 2219235.401 | 1.02E-10 | 1.02E-10 | 1.01E-10 | 1.02E-10 | 1083569.996 | 242291.8529 | 1472206.755 |
|  | mean | 260786.7052 | 1.02E-10 | 1.02E-10 | 1.01E-10 | 1.01E-10 | 225703.8591 | 23799.47495 | 5060.550052 |
|  | median | 247109.3042 | 1.02E-10 | 1.02E-10 | 1.01E-10 | 1.01E-10 | 202596.2197 | 14016.66679 | 6.40E-57 |
|  | stdev | 113140.258 | 1.81E-26 | 1.23E-26 | 1.42E-26 | 3.63E-14 | 99559.98645 | 29724.76944 | 73338.7961 |
| F9 | best | 1.08E-19 | 1.16E-19 | 1.13E-19 | 1.04E-19 | 9.62E-20 | 1.00E-24 | 911.4244623 | 0.15590465 |
|  | worst | 18770598227 | 1.16E-19 | 1.13E-19 | 1.04E-19 | 9.90E-20 | 92699584231 | 1419691418 | 3.69E+11 |
|  | mean | 59781217.28 | 1.16E-19 | 1.13E-19 | 1.04E-19 | 9.62E-20 | 238666361.4 | 5408441.124 | 745438916.6 |
|  | median | 1010.651298 | 1.16E-19 | 1.13E-19 | 1.04E-19 | 9.62E-20 | 4.31E-07 | 1084.6816 | 0.426743166 |
|  | stdev | 1012213432 | 1.44E-35 | 1.69E-35 | 1.87E-35 | 1.25E-22 | 4423166355 | 78721555.39 | 16495993498 |
| F10 | best | 0.000268707 | 0.000271867 | 0.000271857 | 0.000272074 | 0.000271605 | 1305.148346 | 37.97271859 | 0 |
|  | worst | 2365.081755 | 0.000271867 | 0.000271857 | 0.000272074 | 0.000271631 | 2333.501515 | 1211.67998 | 758.1933658 |
|  | mean | 1944.545227 | 0.000271867 | 0.000271857 | 0.000272074 | 0.000271605 | 1869.057115 | 52.82688221 | 3.586552218 |
|  | median | 1927.15985 | 0.000271867 | 0.000271857 | 0.000272074 | 0.000271605 | 1869.71041 | 42.05742491 | 0 |
|  | stdev | 118.8407645 | 4.34E-20 | 3.52E-20 | 4.07E-20 | 1.29E-09 | 346.0828863 | 66.89207141 | 41.44826606 |
| F11 | best | 1.71E-10 | 1.76E-10 | 1.72E-10 | 1.73E-10 | 1.72E-10 | 2394966487 | 4.159217166 | 1.02650854 |
|  | worst | 2594482104 | 1.76E-10 | 1.72E-10 | 1.73E-10 | 1.74E-10 | 2567488472 | 678040718.4 | 429173991.9 |
|  | mean | 975017964 | 1.76E-10 | 1.72E-10 | 1.73E-10 | 1.72E-10 | 2524020304 | 1702811.703 | 1632585.894 |
|  | median | 835513938.7 | 1.76E-10 | 1.72E-10 | 1.73E-10 | 1.72E-10 | 2558711648 | 7.277752467 | 1.102714822 |

|  |  | Compared Algorithms | | | | | | | |
|---|---|---|---|---|---|---|---|---|---|
| **Function** | **Metrics** | **ABC** | **WOA** | **BOA** | **PSO** | **EOSA** | **DE** | **GA** | **HGSO** |
|  | stdev | 310620202.2 | 2.46E-26 | 1.94E-26 | 1.55E-26 | 1.13E-13 | 63720410.11 | 30979999.46 | 22580371.19 |
| F12 | best | 0.00537744 | 0.005388969 | 0.005480457 | 0.005391917 | 0.003344178 | 59.10789733 | 6.976934078 | 9.850801272 |
|  | worst | 106.5810109 | 0.005388969 | 0.005480457 | 0.005391917 | 0.004373574 | 107.3461889 | 63.34963883 | 43.96722626 |
|  | mean | 11.41374051 | 0.005388969 | 0.005480457 | 0.005391917 | 0.003346353 | 81.28584828 | 8.490634765 | 10.05723641 |
|  | median | 3.041798449 | 0.005388969 | 0.005480457 | 0.005391917 | 0.003344178 | 76.95222037 | 7.556693713 | 9.888497941 |
|  | stdev | 19.1491529 | 6.94E-19 | 8.67E-19 | 6.51E-19 | 4.69E-05 | 15.49249829 | 3.739496644 | 1.893231338 |
| F13 | best | 3.65E-10 | 3.74E-10 | 3.66E-10 | 3.73E-10 | 3.74E-10 | 985634985.2 | 17956.26097 | 4950 |
|  | worst | 1384897512 | 3.74E-10 | 3.66E-10 | 3.73E-10 | 3.75E-10 | 1361818654 | 464120523.3 | 340835146.7 |
|  | mean | 889078912.1 | 3.74E-10 | 3.66E-10 | 3.73E-10 | 3.74E-10 | 1254140695 | 1772015.04 | 1336939.589 |
|  | median | 851795363.5 | 3.74E-10 | 3.66E-10 | 3.73E-10 | 3.74E-10 | 1307126803 | 58264.23368 | 4950 |
|  | stdev | 94316100.35 | 3.88E-26 | 5.17E-26 | 5.95E-26 | 5.34E-14 | 129934049.2 | 22235444.11 | 17687769.58 |
| F14 | best | 2.45E-06 | 2.45E-06 | 2.44E-06 | 2.42E-06 | 2.43E-06 | 150745.5792 | 4173.394244 | 0.5 |
|  | worst | 263287.4203 | 2.45E-06 | 2.44E-06 | 2.42E-06 | 2.44E-06 | 262848.6996 | 139559.2807 | 75064.78913 |
|  | mean | 205450.4186 | 2.45E-06 | 2.44E-06 | 2.42E-06 | 2.43E-06 | 207583.4194 | 5816.112605 | 366.1429711 |
|  | median | 201684.2779 | 2.45E-06 | 2.44E-06 | 2.42E-06 | 2.43E-06 | 207654.9189 | 4635.050797 | 0.5 |
|  | stdev | 12935.14014 | 2.33E-22 | 3.39E-22 | 2.33E-22 | 1.64E-09 | 39242.75828 | 7628.700089 | 4110.579697 |
| F15 | best | 2.77E-11 | 2.80E-11 | 2.74E-11 | 2.62E-11 | 2.70E-11 | 1756599198 | 6640180.976 | 1.37E-109 |
|  | worst | 10468218437 | 2.80E-11 | 2.74E-11 | 2.62E-11 | 2.73E-11 | 11188908118 | 5342156920 | 5205941788 |
|  | mean | 5419343203 | 2.80E-11 | 2.74E-11 | 2.62E-11 | 2.70E-11 | 3873865217 | 64605838.05 | 25839038.24 |
|  | median | 5089210478 | 2.80E-11 | 2.74E-11 | 2.62E-11 | 2.70E-11 | 3205180965 | 16145847.47 | 4.70E-55 |
|  | stdev | 849483013.5 | 5.49E-27 | 4.85E-27 | 4.20E-27 | 3.21E-14 | 2157772540 | 295281070.4 | 296296090.2 |
| F16 | best | 0.313183449 | 0.313642786 | 0.313833832 | 0.313841468 | 0.313411811 | 2.953082635 | 2.813194614 | 0 |
|  | worst | 3.06604621 | 0.313642786 | 0.313833832 | 0.313841468 | 0.313489127 | 3.058624131 | 3.049294965 | 3.039113132 |
|  | mean | 2.874483091 | 0.313642786 | 0.313833832 | 0.313841468 | 0.313412095 | 2.979055804 | 2.837486143 | 0.048640529 |
|  | median | 2.858339278 | 0.313642786 | 0.313833832 | 0.313841468 | 0.313411811 | 2.973774936 | 2.828083314 | 0 |
|  | stdev | 0.123478977 | 4.16E-17 | 4.16E-17 | 3.05E-17 | 4.61E-06 | 0.023765099 | 0.034795524 | 0.3529256 |
| F17 | best | 2.74E-11 | 2.79E-11 | 2.77E-11 | 2.74E-11 | 2.78E-11 | 1682987744 | 8124313.895 | 5.04E-111 |
|  | worst | 11088768702 | 2.79E-11 | 2.77E-11 | 2.74E-11 | 2.78E-11 | 10720727065 | 5046542229 | 5836543916 |

| Function | Metrics | Compared Algorithms | | | | | | | |
|---|---|---|---|---|---|---|---|---|---|
| | | ABC | WOA | BOA | PSO | EOSA | DE | GA | HGSO |
| | mean | 5418948553 | 2.79E-11 | 2.77E-11 | 2.74E-11 | 2.78E-11 | 3789413735 | 69537793.4 | 36392422.41 |
| | median | 5064658325 | 2.79E-11 | 2.77E-11 | 2.74E-11 | 2.78E-11 | 3014132286 | 18167648.55 | 4.49E-55 |
| | stdev | 885758421.1 | 4.20E-27 | 4.36E-27 | 4.52E-27 | 3.39E-15 | 2163046329 | 295379702.5 | 357170953.9 |
| F18 | best | 0.065264947 | 0.055534741 | 0.065630095 | 0.065272135 | 0.046648475 | 11.29812226 | 4.736804364 | 0 |
| | worst | 11.507642 | 0.065329978 | 0.065630095 | 0.065272135 | 0.058911739 | 11.44580276 | 10.09038934 | 8.480262936 |
| | mean | 2.862967765 | 0.057169632 | 0.065630095 | 0.065272135 | 0.046680024 | 11.40128365 | 5.190692848 | 0.114397103 |
| | median | 1.559911469 | 0.055534741 | 0.065630095 | 0.065272135 | 0.046648475 | 11.44302638 | 4.917419975 | 0 |
| | stdev | 2.575656552 | 0.003463235 | 1.04E-17 | 7.63E-18 | 0.000605181 | 0.064592634 | 0.739157175 | 0.754170922 |
| F19 | best | 0.018649694 | 0.018759907 | 0.018764486 | 0.018698899 | 0.018690559 | 45.32931639 | 42.22016808 | 0 |
| | worst | 46.70160803 | 0.018759907 | 0.018764486 | 0.018698899 | 0.018746242 | 46.47403905 | 45.65285146 | 46.27336925 |
| | mean | 35.5326682 | 0.018759907 | 0.018764486 | 0.018698899 | 0.018690719 | 45.65823826 | 42.34684559 | 1.694447863 |
| | median | 34.73786127 | 0.018759907 | 0.018764486 | 0.018698899 | 0.018690559 | 45.65811462 | 42.33630398 | 1.134235292 |
| | stdev | 2.631279754 | 2.60E-18 | 2.26E-18 | 1.91E-18 | 2.87E-06 | 0.31817398 | 0.272793571 | 5.762532227 |
| F20 | best | 0.000250786 | 0.000246696 | 0.000249866 | 0.000252195 | 0.000246298 | 13.25336872 | 40.41662168 | 11.51810893 |
| | worst | 1477.956124 | 0.000246696 | 0.000249866 | 0.000252195 | 0.000247986 | 1437.269586 | 820.6332779 | 669.1654077 |
| | mean | 106.6199225 | 0.000246696 | 0.000249866 | 0.000252195 | 0.000246305 | 225.3094727 | 57.42482498 | 29.56622817 |
| | median | 14.32225309 | 0.000246696 | 0.000249866 | 0.000252195 | 0.000246298 | 64.6037593 | 46.02516801 | 19.93840466 |
| | stDev | 235.2224648 | 2.98E-20 | 4.61E-20 | 3.39E-20 | 1.05E-07 | 322.074818 | 49.11711471 | 42.50885548 |
| F21 | best | 0.001479242 | 0.001517382 | 0.001515007 | 0.001483363 | 0.001456843 | 6.978693002 | 20.75932802 | 0.75338742 |
| | worst | 323.037448 | 0.001517382 | 0.001515007 | 0.001483363 | 0.001525863 | 315.7329145 | 185.6204308 | 148.4719403 |
| | mean | 20.33662271 | 0.001517382 | 0.001515007 | 0.001483363 | 0.001457027 | 57.82178336 | 25.69890585 | 1.674645626 |
| | median | 2.67861904 | 0.001517382 | 0.001515007 | 0.001483363 | 0.001456843 | 28.30660489 | 22.13469236 | 0.753953307 |
| | stdev | 48.64424139 | 2.49E-19 | 1.73E-19 | 1.63E-19 | 3.40E-06 | 69.53001349 | 11.49308344 | 9.058474116 |
| F22 | best | 0.000267348 | 0.000268108 | 0.000264328 | 0.000267984 | 0.000122824 | 978.4644232 | 0.098573292 | 4.79E-05 |
| | worst | 1681.083806 | 0.000268108 | 0.000264328 | 0.000267984 | 0.00019473 | 1747.934015 | 561.5855192 | 247.2965395 |
| | mean | 115.8899526 | 0.000268108 | 0.000264328 | 0.000267984 | 0.000122984 | 1447.842362 | 2.207505659 | 0.73019635 |
| | median | 2.416271301 | 0.000268108 | 0.000264328 | 0.000267984 | 0.000122824 | 1484.787158 | 0.135816598 | 8.23E-05 |
| | stdev | 300.0316907 | 4.61E-20 | 3.39E-20 | 4.07E-20 | 3.33E-06 | 279.6249961 | 26.86476596 | 11.68303402 |

|  |  | Compared Algorithms | | | | | | | |
|---|---|---|---|---|---|---|---|---|---|
| Function | Metrics | ABC | WOA | BOA | PSO | EOSA | DE | GA | HGSO |
| F23 | best | -302.5008149 | -262.1755865 | -182.7543715 | -294.1230158 | -16.54091663 | -295.6844661 | 0.000444 | -270.8437899 |
|  | worst | -2.431610245 | -226.2954225 | -172.0744886 | -204.9868298 | -2.618649088 | -221.660774 | 0.645516 | -220.8575849 |
|  | mean | -295.8858845 | -259.1191967 | -182.7082986 | -284.8981188 | -16.45742878 | -290.057282 | 0.016006 | -266.4443466 |
|  | median | -301.8715247 | -260.0368176 | -182.7543715 | -292.770422 | -16.54091663 | -293.8833165 | 0.000444 | -268.0098591 |
|  | stdev | 19.05436824 | 4.240456788 | 0.678269937 | 15.03776577 | 1.074758903 | 9.805001586 | 0.079374 | 6.182581047 |
| F25 | best | 1.73E-05 | 2.02E-05 | 2.19E-05 | 1.99E-05 | 2.22E-05 | 2.05E-29 | 0.005818419 | 2.98E-112 |
|  | worst | 24.75787505 | 2.02E-05 | 2.19E-05 | 1.99E-05 | 2.25E-05 | 15.6011816 | 5.764924377 | 14.05993536 |
|  | mean | 0.336241769 | 2.02E-05 | 2.19E-05 | 1.99E-05 | 2.22E-05 | 0.128138027 | 0.035160231 | 0.036388951 |
|  | median | 0.004759136 | 2.02E-05 | 2.19E-05 | 1.99E-05 | 2.22E-05 | 1.17E-14 | 0.007171468 | 4.30E-51 |
|  | stdev | 1.716830508 | 3.05E-21 | 4.07E-21 | 2.71E-21 | 5.48E-08 | 1.159489759 | 0.298540229 | 0.653830959 |
| F26 | best | 1.37E-10 | 1.36E-10 | 1.39E-10 | 1.39E-10 | 1.38E-10 | 2908169157 | 36768.89195 | 2.36E-07 |
|  | worst | 3850449321 | 1.36E-10 | 1.39E-10 | 1.39E-10 | 1.39E-10 | 3685252577 | 1215212126 | 526426653.9 |
|  | mean | 2393795142 | 1.36E-10 | 1.39E-10 | 1.39E-10 | 1.38E-10 | 3376494490 | 4715971.794 | 1701158.819 |
|  | median | 2309813185 | 1.36E-10 | 1.39E-10 | 1.39E-10 | 1.38E-10 | 3399311165 | 165504.1559 | 2.36E-07 |
|  | stdev | 243210513.4 | 1.42E-26 | 1.94E-26 | 1.42E-26 | 3.00E-14 | 300829422.2 | 58549154.5 | 25984261.2 |
| F27 | best | 0.00047527 | 0.00047985 | 0.000471891 | 0.000473586 | 0.000448786 | 1347.380357 | 749.9920473 | 0 |
|  | worst | 1603.785958 | 0.00047985 | 0.000471891 | 0.000473586 | 0.000470662 | 1602.675545 | 1280.232203 | 1184.58447 |
|  | mean | 451.6267178 | 0.00047985 | 0.000471891 | 0.000473586 | 0.000448847 | 1440.929879 | 772.4357583 | 11.72060634 |
|  | median | 323.4669501 | 0.00047985 | 0.000471891 | 0.000473586 | 0.000448786 | 1418.535827 | 759.7515676 | 0 |
|  | stdev | 270.0398663 | 8.40E-20 | 5.15E-20 | 5.96E-20 | 1.10E-06 | 79.40879312 | 44.56932118 | 99.55250013 |
| F28 | best | 1.07E-08 | 9.70E-09 | 9.31E-09 | 1.00E-08 | 9.25E-09 | 466932.353 | 12844.73759 | 1.67E-114 |
|  | worst | 1022160.738 | 9.70E-09 | 9.31E-09 | 1.00E-08 | 9.93E-09 | 1011320.133 | 495929.0155 | 884512.6177 |
|  | mean | 373004.6264 | 9.70E-09 | 9.31E-09 | 1.00E-08 | 9.25E-09 | 561177.5755 | 20051.98473 | 5008.283742 |
|  | median | 344810.3587 | 9.70E-09 | 9.31E-09 | 1.00E-08 | 9.25E-09 | 537590.8983 | 15053.61311 | 2.68E-54 |
|  | stdev | 81563.46341 | 1.24E-24 | 1.36E-24 | 1.82E-24 | 3.71E-11 | 106070.0836 | 28150.07853 | 49908.74417 |
| F29 | best | 4.59E-10 | 4.56E-10 | 4.50E-10 | 4.62E-10 | 4.51E-10 | 921421360.4 | 16533.59328 | 98.86771563 |
|  | worst | 1085749209 | 4.56E-10 | 4.50E-10 | 4.62E-10 | 4.57E-10 | 1105359990 | 365586773.6 | 237130959.5 |
|  | mean | 391773892 | 4.56E-10 | 4.50E-10 | 4.62E-10 | 4.51E-10 | 1041990575 | 1472601.153 | 746936.1114 |

|          |         | **Compared Algorithms** | | | | | | | |
| Function | Metrics | ABC | WOA | BOA | PSO | EOSA | DE | GA | HGSO |
|---|---|---|---|---|---|---|---|---|---|
|     | median | 327193480.7 | 4.56E-10 | 4.50E-10 | 4.62E-10 | 4.51E-10 | 1048794876 | 58273.67518 | 98.88822056 |
|     | stdev  | 141540445.4 | 6.98E-26 | 6.98E-26 | 9.31E-26 | 3.24E-13 | 68984880.73 | 17684842.71 | 11579159.04 |
| F30 | best   | -8836.361886 | -41041.09361 | -82.284136 | -8.37E+13 | -1.77E+23 | -10720.24772 | 0.026042 | -19627.1811 |
|     | worst  | -0.06884986 | -14838.49714 | -55.03621215 | -4236.486476 | -0.073719104 | -5231.572209 | 0.116496 | -4632.384271 |
|     | mean   | -8466.742315 | -40720.29286 | -82.15043677 | -4.29E+13 | -1.75E+23 | -9668.267461 | 0.051781 | -15952.69912 |
|     | median | -8755.05295 | -41028.05197 | -82.284136 | -5.09E+13 | -1.77E+23 | -9966.438769 | 0.039345 | -16808.49531 |
|     | stdev  | 735.9984565 | 1938.702109 | 1.754877788 | 3.92E+13 | 1.76E+22 | 1097.48294 | 0.035806 | 3447.147118 |
| F31 | best   | 9.28E-09 | 1.08E-08 | 1.06E-08 | 1.04E-08 | 8.71E-09 | 466276.9769 | 13040.44756 | 1.04E-116 |
|     | worst  | 1183271.593 | 1.08E-08 | 1.06E-08 | 1.04E-08 | 8.99E-09 | 1097796.317 | 507957.0629 | 922710.1499 |
|     | mean   | 403966.8473 | 1.08E-08 | 1.06E-08 | 1.04E-08 | 8.71E-09 | 568102.224 | 20382.15347 | 4580.700889 |
|     | median | 369094.2037 | 1.08E-08 | 1.06E-08 | 1.04E-08 | 8.71E-09 | 547682.4572 | 15390.81415 | 6.64E-59 |
|     | stdev  | 95470.2388 | 1.08E-24 | 1.74E-24 | 1.86E-24 | 1.61E-11 | 109961.5849 | 28563.43898 | 50858.65247 |
| F32 | best   | 7.03E-167 | 1.69E-166 | 3.43E-166 | 2.37E-165 | 1.07E-166 | 4.99E+133 | 445.0616521 | 1.73E-58 |
|     | worst  | 8.34E+147 | 1.69E-166 | 3.43E-166 | 2.37E-165 | 4.44E-166 | 2.60E+146 | 3.58E+143 | 4.82E+135 |
|     | mean   | 1.68E+145 | 1.69E-166 | 3.43E-166 | 2.37E-165 | 1.26E-166 | 6.15E+144 | 7.26E+140 | 9.64E+132 |
|     | median | 1.40E+106 | 1.69E-166 | 3.43E-166 | 2.37E-165 | 1.07E-166 | 7.01E+138 | 447.2618729 | 6.20E-31 |
|     | stdev  | 3.73E+146 | 0 | 0 | 0 | 0 | 3.61E+145 | 1.60E+142 | 2.15E+134 |
| F33 | best   | 0.010001246 | 0.010001466 | 0.010000903 | 0.010000615 | 0.009960159 | 94.68547115 | 17.33513207 | 1.52E-57 |
|     | worst  | 94.81833964 | 0.010001466 | 0.010000903 | 0.010000615 | 0.010001165 | 94.68547115 | 81.02072384 | 57.37997572 |
|     | mean   | 87.26062108 | 0.010001466 | 0.010000903 | 0.010000615 | 0.009960264 | 94.68547115 | 19.58660243 | 0.520898377 |
|     | median | 86.78228412 | 0.010001466 | 0.010000903 | 0.010000615 | 0.009960159 | 94.68547115 | 18.15536627 | 4.91E-27 |
|     | stdev  | 4.179693275 | 7.81E-19 | 7.81E-19 | 1.39E-18 | 1.95E-06 | 8.53E-15 | 5.107576845 | 4.002551529 |
| F34 | best   | 2.39E-06 | 2.43E-06 | 2.49E-06 | 2.44E-06 | 2.42E-06 | 146564.3791 | 4151.903716 | 2.56E-92 |
|     | worst  | 257895.3838 | 2.43E-06 | 2.49E-06 | 2.44E-06 | 2.43E-06 | 257656.427 | 134925.0854 | 88039.69375 |
|     | mean   | 208128.671 | 2.43E-06 | 2.49E-06 | 2.44E-06 | 2.42E-06 | 200833.5123 | 5709.442348 | 687.5080256 |
|     | median | 204667.8202 | 2.43E-06 | 2.49E-06 | 2.44E-06 | 2.42E-06 | 204872.6698 | 4541.823055 | 4.43E-29 |
|     | stdev  | 13021.70178 | 1.91E-22 | 3.39E-22 | 2.33E-22 | 1.87E-10 | 40515.11285 | 7456.617438 | 5515.873891 |
| F35 | best   | 2.47E-06 | 2.43E-06 | 2.45E-06 | 2.44E-06 | 2.43E-06 | 149354.2611 | 4296.604798 | 19.93229264 |

|   | | Compared Algorithms | | | | | | | |
|---|---|---|---|---|---|---|---|---|---|
| Function | Metrics | ABC | WOA | BOA | PSO | EOSA | DE | GA | HGSO |
|  | worst | 257643.1306 | 2.43E-06 | 2.45E-06 | 2.44E-06 | 2.44E-06 | 261762.2494 | 138652.6906 | 91297.1041 |
|  | mean | 205778.2306 | 2.43E-06 | 2.45E-06 | 2.44E-06 | 2.43E-06 | 204473.9907 | 5798.078168 | 401.736332 |
|  | median | 201888.0566 | 2.43E-06 | 2.45E-06 | 2.44E-06 | 2.43E-06 | 204296.1293 | 4439.333834 | 21.13665172 |
|  | stdev | 13031.29588 | 3.18E-22 | 3.18E-22 | 2.33E-22 | 5.73E-10 | 39642.75925 | 7590.203072 | 4823.963514 |
| F36 | best | 4.66E-06 | 4.61E-06 | 4.71E-06 | 4.72E-06 | 4.61E-06 | 66187.97685 | 1582.956209 | 1.63E-119 |
|  | worst | 125317.1139 | 4.61E-06 | 4.71E-06 | 4.72E-06 | 4.66E-06 | 124112.1235 | 66662.76425 | 39912.02993 |
|  | mean | 27848.95235 | 4.61E-06 | 4.71E-06 | 4.72E-06 | 4.61E-06 | 92944.77725 | 2388.574257 | 153.7211311 |
|  | median | 18307.7666 | 4.61E-06 | 4.71E-06 | 4.72E-06 | 4.61E-06 | 90158.00805 | 1737.54854 | 2.68E-60 |
|  | stdev | 20873.64614 | 4.24E-22 | 7.62E-22 | 5.93E-22 | 2.90E-09 | 20130.48263 | 3711.651491 | 2089.954545 |
| F37 | best | 0.000305122 | 0.023335766 | 0.024544119 | 0.024575999 | 0.011110877 | 14.4645922 | 0.409466028 | 6.78E-122 |
|  | worst | 25.8151169 | 0.024556773 | 0.024544119 | 0.024575999 | 0.016303029 | 25.64204569 | 13.6419851 | 11.58066632 |
|  | mean | 3.086996871 | 0.023597731 | 0.024544119 | 0.024575999 | 0.011125287 | 19.72266104 | 0.56836581 | 0.060472667 |
|  | median | 0.118511189 | 0.023335766 | 0.024544119 | 0.024575999 | 0.011110877 | 18.95500975 | 0.447697234 | 3.98E-60 |
|  | stdev | 5.923666804 | 0.000498763 | 3.64E-18 | 3.64E-18 | 0.000258258 | 3.843309197 | 0.757549897 | 0.658790887 |
| F38 | best | 8.34E-200 | 4.53E-200 | 3.09E-200 | 2.58E-200 | 1.58E-200 | 1.03E+153 | 1.50E+61 | 2.11E-62 |
|  | worst | 1.21E+178 | 4.53E-200 | 3.09E-200 | 2.58E-200 | 2.18E-200 | 2.36E+175 | 1.89E+172 | 4.92E+167 |
|  | mean | 2.42E+175 | 4.53E-200 | 3.09E-200 | 2.58E-200 | 1.58E-200 | 1.34E+173 | 3.78E+169 | 9.84E+164 |
|  | median | 1.22E+162 | 4.53E-200 | 3.09E-200 | 2.58E-200 | 1.58E-200 | 1.35E+160 | 3.27E+81 | 1.96E-34 |
|  | stdev | 0 | 0 | 0 | 0 | 0 | 0 | 0 | 0 |
| F39 | best | 2.46E-12 | 2.48E-12 | 2.47E-12 | 2.46E-12 | 2.46E-12 | 1.49E+11 | 4087014600 | 78062850.06 |
|  | worst | 2.60533E+11 | 2.48E-12 | 2.47E-12 | 2.46E-12 | 2.47E-12 | 2.61E+11 | 1.35E+11 | 75834606829 |
|  | mean | 2.02724E+11 | 2.48E-12 | 2.47E-12 | 2.46E-12 | 2.46E-12 | 2.05E+11 | 5631425626 | 752360824.7 |
|  | median | 1.99043E+11 | 2.48E-12 | 2.47E-12 | 2.46E-12 | 2.46E-12 | 2.00E+11 | 4422413197 | 83482644.88 |
|  | stdev | 12748137431 | 3.43E-28 | 3.23E-28 | 3.03E-28 | 2.02E-16 | 39244378992 | 7381450963 | 5182164328 |
| F40 | best | 3.01E-70 | 3.60E-70 | 2.93E-70 | 2.61E-70 | 2.02E-70 | 8.61E+48 | 1200.2428 | 1200.00029 |
|  | worst | 4.50E+54 | 3.60E-70 | 2.93E-70 | 2.61E-70 | 2.53E-70 | 7.42E+54 | 1.37E+51 | 2.72E+54 |
|  | mean | 1.09E+52 | 3.60E-70 | 2.93E-70 | 2.61E-70 | 2.02E-70 | 5.52E+53 | 5.37E+48 | 1.08E+52 |
|  | median | 3.36E+45 | 3.60E-70 | 2.93E-70 | 2.61E-70 | 2.02E-70 | 9.18E+50 | 1201.064629 | 1200.000481 |

|          |         | Compared Algorithms |             |             |             |             |             |             |             |
|----------|---------|---------------------|-------------|-------------|-------------|-------------|-------------|-------------|-------------|
| Function | Metrics | ABC                 | WOA         | BOA         | PSO         | EOSA        | DE          | GA          | HGSO        |
|          | stdev   | 2.07E+53            | 3.58E-86    | 6.64E-86    | 4.02E-86    | 2.66E-72    | 1.69E+54    | 8.19E+49    | 1.71E+53    |
| F41      | best    | 5.69E-28            | 5.72E-28    | 5.76E-28    | 5.76E-28    | 5.53E-28    | 3.91E+26    | 1.13E+27    | 5.69E+26    |
|          | worst   | 1.09E+27            | 5.72E-28    | 5.76E-28    | 5.76E-28    | 5.65E-28    | 1.04E+27    | 1.13E+27    | 9.89E+26    |
|          | mean    | 9.31E+26            | 5.72E-28    | 5.76E-28    | 5.76E-28    | 5.53E-28    | 5.32E+26    | 1.13E+27    | 6.56E+26    |
|          | median  | 9.26E+26            | 5.72E-28    | 5.76E-28    | 5.76E-28    | 5.53E-28    | 5.00E+26    | 1.13E+27    | 6.32E+26    |
|          | stdev   | 5.15E+25            | 7.17E-44    | 7.17E-44    | 5.83E-44    | 5.93E-31    | 1.31E+26    | 1.61E+24    | 7.69E+25    |
| F42      | best    | 4.27E-06            | 4.29E-06    | 4.26E-06    | 4.16E-06    | 4.09E-06    | 31976.73516 | 1188.933345 | 421.3277663 |
|          | worst   | 108535.2566         | 4.29E-06    | 4.26E-06    | 4.16E-06    | 4.17E-06    | 108929.7038 | 48344.70056 | 32891.91673 |
|          | mean    | 62929.59696         | 4.29E-06    | 4.26E-06    | 4.16E-06    | 4.09E-06    | 61436.09831 | 1620.118778 | 559.237889  |
|          | median  | 60073.02152         | 4.29E-06    | 4.26E-06    | 4.16E-06    | 4.09E-06    | 57366.81193 | 1310.196265 | 421.5626519 |
|          | stdev   | 7334.315184         | 5.08E-22    | 5.08E-22    | 4.24E-22    | 5.31E-09    | 25706.21856 | 2466.571014 | 1710.546195 |
| F43      | best    | 0.000330323         | 0.000332408 | 0.000333684 | 0.000333589 | 0.00033233  | 2266.842103 | 1643.429176 | 942.528339  |
|          | worst   | 2479.036365         | 0.000332408 | 0.000333684 | 0.000333589 | 0.000333577 | 2493.325685 | 2169.822427 | 2069.438537 |
|          | mean    | 1909.413375         | 0.000332408 | 0.000333684 | 0.000333589 | 0.000332333 | 2359.039142 | 1679.482761 | 964.0878109 |
|          | median  | 1848.634581         | 0.000332408 | 0.000333684 | 0.000333589 | 0.00033233  | 2344.668872 | 1670.803483 | 944.5733014 |
|          | stdev   | 157.8611421         | 3.79E-20    | 4.88E-20    | 3.79E-20    | 5.80E-08    | 66.81285479 | 46.64354787 | 104.1875831 |
| F44      | best    | 2.09E-29            | 1.78E-29    | 1.89E-29    | 2.03E-29    | 1.79E-29    | 438320.8487 | 110941.4238 | 592240.2576 |
|          | worst   | 2.76E+24            | 1.78E-29    | 1.89E-29    | 2.03E-29    | 1.93E-29    | 1.10E+24    | 1.04E+24    | 1.92E+23    |
|          | mean    | 1.43E+22            | 1.78E-29    | 1.89E-29    | 2.03E-29    | 1.79E-29    | 3.47E+21    | 3.14E+21    | 6.82E+20    |
|          | median  | 4.57E+18            | 1.78E-29    | 1.89E-29    | 2.03E-29    | 1.79E-29    | 463752.3183 | 137938.6535 | 614132.7943 |
|          | stdev   | 1.59E+23            | 2.24E-45    | 3.01E-45    | 2.59E-45    | 6.78E-32    | 5.64E+22    | 5.05E+22    | 1.03E+22    |
| F45      | best    | 0.305715386         | 0.276970281 | 0.308726297 | 0.305443053 | 0.30646382  | 2.713949275 | 2.704585339 | 0           |
|          | worst   | 2.834155251         | 0.307895716 | 0.308726297 | 0.305443053 | 0.307290867 | 2.830118506 | 2.794061234 | 2.6140882   |
|          | mean    | 1.740456859         | 0.281499603 | 0.308726297 | 0.305443053 | 0.306468731 | 2.735885851 | 2.704932219 | 0.045569177 |
|          | median  | 1.619931442         | 0.276970281 | 0.308726297 | 0.305443053 | 0.30646382  | 2.723019587 | 2.704585339 | 0           |
|          | stdev   | 0.264814711         | 0.010094625 | 3.61E-17    | 3.61E-17    | 6.02E-05    | 0.029287729 | 0.005105112 | 0.286856855 |
| F46      | best    | 2.46E-05            | 2.46E-05    | 2.44E-05    | 2.43E-05    | 2.44E-05    | 14557.36666 | 409.0949855 | 8.28E-108   |
|          | worst   | 26682.39023         | 2.46E-05    | 2.44E-05    | 2.43E-05    | 2.44E-05    | 26064.68463 | 13842.05832 | 9942.048708 |

|  |  | Compared Algorithms | | | | | | | |
|---|---|---|---|---|---|---|---|---|---|
| **Function** | **Metrics** | **ABC** | **WOA** | **BOA** | **PSO** | **EOSA** | **DE** | **GA** | **HGSO** |
|  | mean | 20454.91167 | 2.46E-05 | 2.44E-05 | 2.43E-05 | 2.44E-05 | 20192.37282 | 565.2503799 | 57.57076258 |
|  | median | 20071.41036 | 2.46E-05 | 2.44E-05 | 2.43E-05 | 2.44E-05 | 19688.65321 | 451.293936 | 5.07E-48 |
|  | stdev | 1324.806232 | 2.71E-21 | 3.22E-21 | 3.22E-21 | 4.47E-09 | 3845.341638 | 751.0738746 | 599.829055 |
| F47 | best | 0.005888844 | 0.00587983 | 0.005866289 | 0.005919446 | 0.005858814 | 99.71266277 | 14.12662111 | 0 |
|  | worst | 131.4206624 | 0.00587983 | 0.005866289 | 0.005919446 | 0.005890008 | 129.9725436 | 98.5658473 | 68.10191817 |
|  | mean | 31.30562957 | 0.00587983 | 0.005866289 | 0.005919446 | 0.005858927 | 115.3842895 | 16.63768708 | 0.393527918 |
|  | median | 7.982565255 | 0.00587983 | 0.005866289 | 0.005919446 | 0.005858814 | 114.8670662 | 14.94988815 | 0 |
|  | stdev | 39.53996915 | 6.07E-19 | 7.81E-19 | 5.64E-19 | 1.84E-06 | 10.10326971 | 6.603864967 | 4.244564029 |

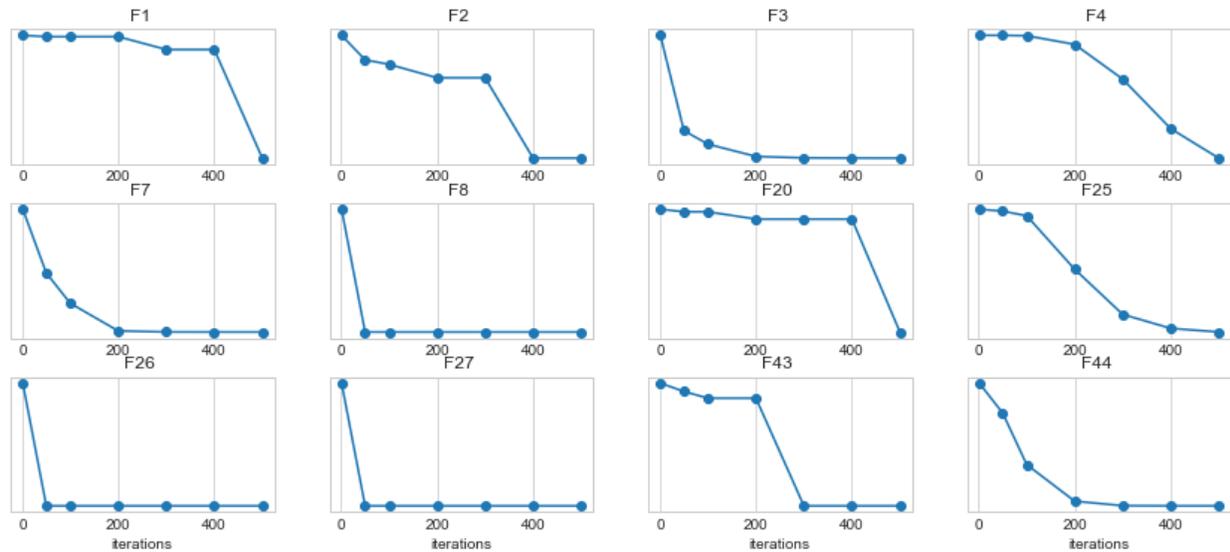

**Figure 4:** Convergent curves of EOSA on some selected standard benchmark functions over 1, 50, 100, 200, 300, 400, and 500 epochs

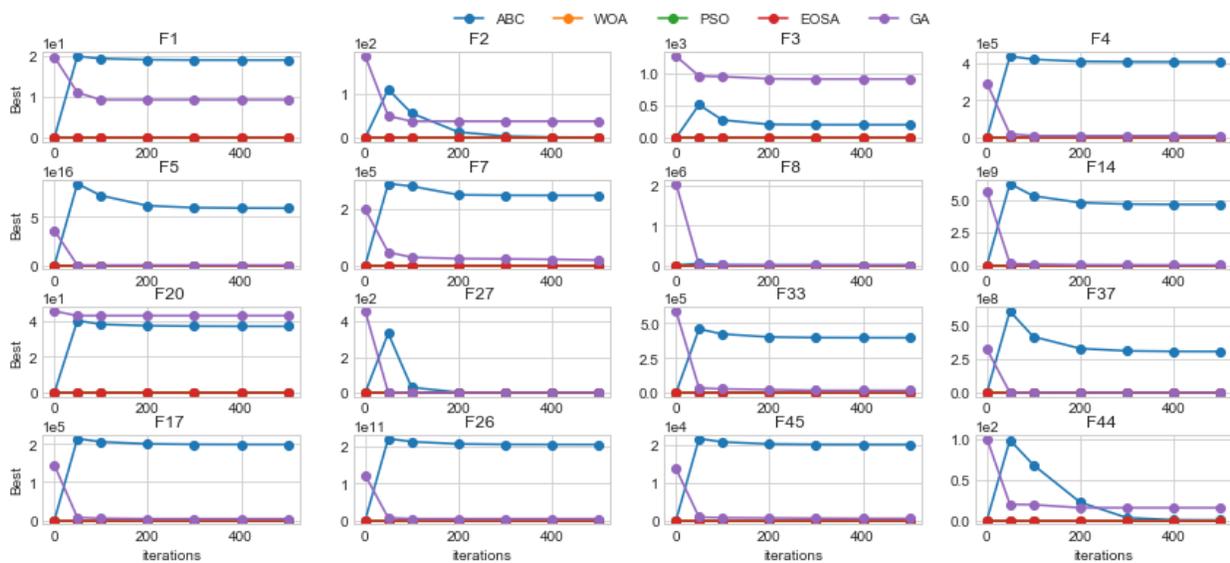

**Figure 5:** Convergent curves of EOSA and related optimization algorithms in some selected standard benchmark functions

To confirm the good performance and superiority of the EOSA, in Figure 4, the convergence curves of the hostiory of solutions are graphed. The benchmark functions F1, F2, F3, F4, F7, F8, F20, F25, F26, F27, F43, and F45. We observed that in all cases, the curve of the plots descended appreciably. The implication is that the EOSA algorithm is a very competitive optimization algorithm that can discover optimum solutions both in the exploration and exploitation operations. To demonstrate the superiority of EOSA, when its convergence curve is plotted against those of state-of-the-art algorithms, we found that its solutions are significant.

The convergence of EOSA using the benchmark functions as compared ABC, WOA, PSO, and GA was graphed and illustrated in Figure 5. In most cases, the figures showed that the best values for all the algorithms were descending from their initial peak values to a lower value as the training improves over some epoch. The graphing was achieved by obtaining the best values for each case of EOSA, ABC, WOA, PSO, and GA at 1, 50, 100, 200, 300, 400, 500 iterations points. The figure confirms that the best values for GA are often larger than the others except in a few cases were ABC also has some large values. However, it is clear that ABC, WOA, PSO, and the proposed EOSA do not just have low value for the best cases but only drop in value for small fractions across all the iterations. This is why their rise and fall in relation to their curves was not so pronounced in comparison with that of GA and sometimes ABC.

Perspective views, alongside the search history, of some selected functions, are shown in Figures 6 and 7. Again, to compare the performance of EOSA with other similar optimization algorithms, the illustrations in the figures were graphed with all the selected algorithms. The outcome for F1function showed that EOSA appears to converge its points more closely compared with ABC while WOA showed only a point. However, in F3 and F34, PSO and GA respectively converged their point more closely than EOSA. Meanwhile, as in the F1 function, we found that EOSA converges its points more closely than BOA and ABC.

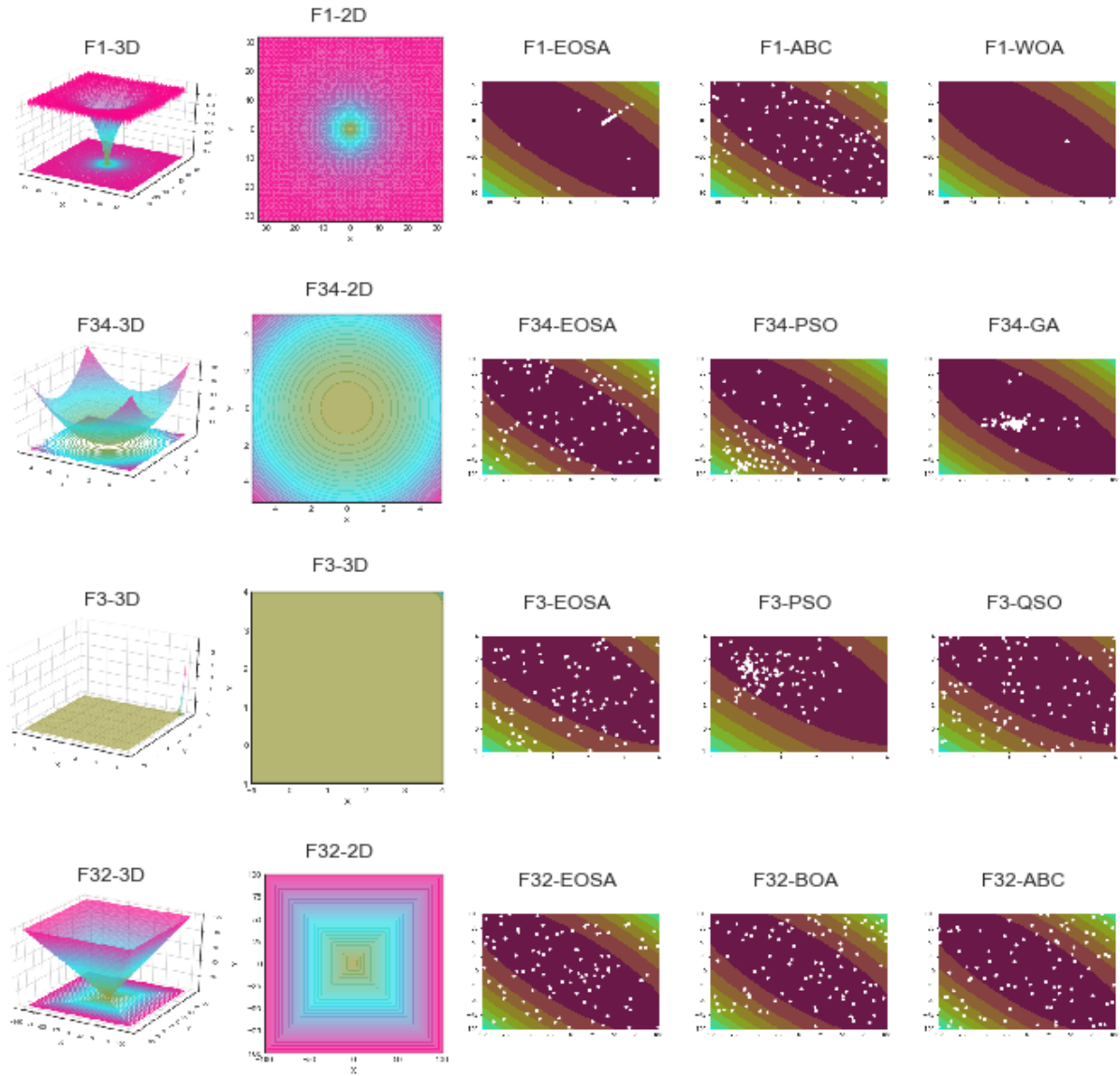

**Figure 6:** A 3D and 2D perspective view for functions F1, F3, F32, and F34 to 26 and their corresponding search history for EOSA and related optimization algorithms

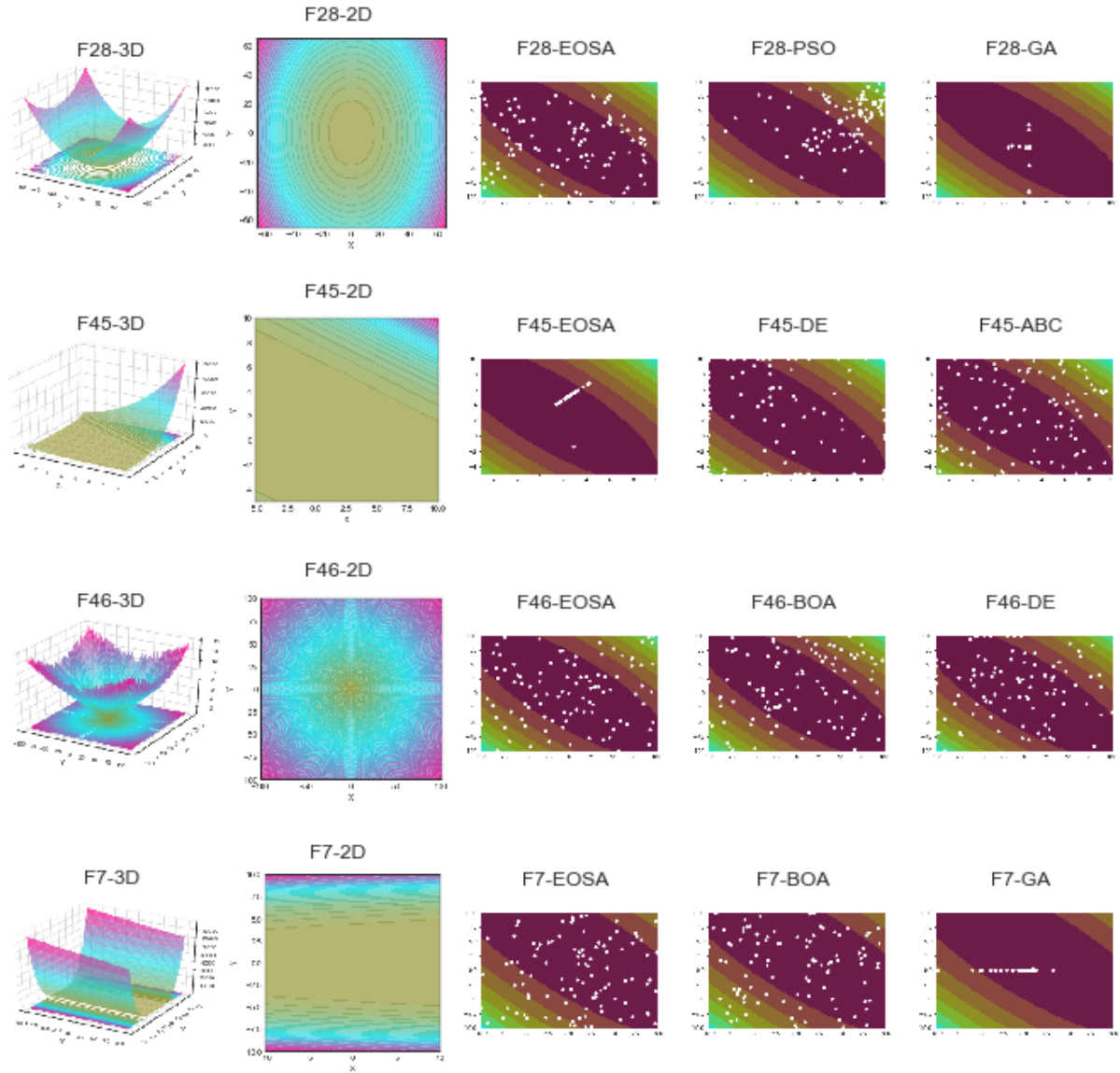

**Figure 7:** A 3D and 2D perspective view for functions F7, F28, F45, and F46 and their corresponding search history for EOSA and related optimization algorithms

The second illustration is shown in Figure 6 that only in F7 and F28 was GA able to cluster its points more closely compared with EOSA, PSO, BOA, DE, and ABC. While PSO attempts to achieve a clearer cluster in F28, EOSA also performs well in clustering of points or solutions in F7, F28, F45, and F46.

### 5.3 Evaluation of EOSA on Constrained CEC Benchmark functions

The result of experimentation with constrained functions of CEC was also collected and reported in this subsection. The CEC functions consist of fourteen (14) functions, and we further experimented with thirty (30) hybrids of the 14 CEC functions. The derivation for the hybrid functions is listed in Table 5. Functions C1-C8 and C10 are shifted versions of their corresponding CEC functions, while C9 and C11-C16 are shift-rotate versions of their corresponding CEC functions. Functions C17-C22 represent shift versions of a combination of some CEC functions, while those of

C23-C30 are hybrids of predefined CEC hybrids. The total number of CEC-based functions used for the experiments is forty-four (44).

**Table 5**: Listing of thirty (30) hybrids of CEC functions applied to the proposed EOSA algorithm where shift is (S), and Shift rotate (SF)

| Function | Hybrid CEC Functions | Function | Hybrid CEC Functions |
|---|---|---|---|
| C1 | S CEC01 | C16 | SR CEC14 |
| C2 | S CEC02 | C17 | S [CEC09, CEC08, CEC01] |
| C3 | S CEC03 | C18 | S [CEC02, CEC12, CEC08] |
| C4 | S CEC04 | C19 | S [CEC07, CEC06, CEC04, CEC14] |
| C5 | S CEC05 | C20 | S [CEC12, CEC03, CEC13, CEC08] |
| C6 | S CEC06 | C21 | S [CEC14, CEC12, CEC04, CEC09, CEC01] |
| C7 | S CEC07 | C22 | S [CEC10, CEC11, CEC13, CEC09, CEC05] |
| C8 | S CEC08 | C23 | S(1,2,3,4,5) [C04, C01, C02, C03, C01] |
| C9 | SR CEC08 | C24 | S(1,2,3) [C10, C09, C14] |
| C10 | S CEC09 | C25 | S(1,2,3) [C11, C09, C01] |
| C11 | SR CEC09 | C26 | S(1,2,3,4,5) [C11, C13, C01, C06, C07] |
| C12 | SR CEC10 | C27 | S(1,2,3,4,5) [C14, C09, C11, C06, C01] |
| C13 | SR CEC11 | C28 | S(1,2,3,4,5) [C15, C13, C13, C11, C16, C1] |
| C14 | SR CEC12 | C29 | S(4,5,6) [C17, C18, C19] |
| C15 | SR CEC13 | C30 | S(1,2,3) [C20, C21, C22] |

Tables 6, 7, 8, 9 and 10 outline the outcome of experimenting with the forty-four (44) CEC functions regarding the best values, mean values, standard deviation, worst values, and median values, respectively. The results compare the performances of ABC, WOA, BOA PSO, DE, GA, and HGSO with EOSA based on the categories of outcome each table outlines. Regarding the best values, the result showed that EOSA outperformed all the related algorithms based on the CEC01, CEC03, CEC05-06, C3-4, C7-8, C15-16, C19-28, and C30 functions. Whereas it failed when compared with those related algorithms based on CEC07-CEC11, C2, C17-18, and C29 functions, we found out that it competed with the optimization algorithms based on CEC02 CEC04, CEC12-14, C1, C6, and C9-14 functions.

**Table 6:** Comparison of best values for EOSA with ABC, WOA, BOA PSO, DE, GA, and HGSO metaheuristic algorithms using the CEC functions over 20 runs and 100 population size

|  | ABC | WOA | BOA | PSO | EOSA | DE | GA | HGSO |
|---|---|---|---|---|---|---|---|---|
| **CEC_F1** | 2.78E-11 | 2.78E-11 | 2.75E-11 | 2.78E-11 | 2.75E-11 | 1.67E+09 | 6500451 | 1.22E-111 |
| **CEC_F2** | 2.49E-12 | 2.45E-12 | 2.48E-12 | 2.49E-12 | 2.48E-12 | 1.43E+11 | 4.17E+09 | 2.15E-104 |
| **CEC_F3** | 1.02E-10 | 1.02E-10 | 1.02E-10 | 1.03E-10 | 1.01E-10 | 150261.7 | 8666.065 | 2.06E-112 |
| **CEC_F4** | 3.68E-12 | 3.70E-12 | 3.73E-12 | 3.73E-12 | 3.71E-12 | 1.09E+11 | 1099091 | 98.82988 |
| **CEC_F5** | 0.045719 | 0.045711 | 0.045704 | 0.045704 | 0.045669 | 20.00005 | 18.25292 | 4.44E-16 |
| **CEC_F6** | 0.005831 | 0 | 0.005899 | 0.005873 | 0.005742 | 0 | 116.5891 | 0 |
| **CEC_F7** | 0.009643 | 0.009774 | 0.009701 | 0.009743 | 0.009735 | 41.81979 | 2.200358 | 0 |
| **CEC_F8** | 2.44E-06 | 2.40E-06 | 2.45E-06 | 2.44E-06 | 2.43E-06 | 136298.4 | 5267.792 | 0 |
| **CEC_F9** | 2.92E-05 | 2.91E-05 | 2.92E-05 | 2.94E-05 | 2.94E-05 | 14531.82 | 564.2694 | 0.001273 |
| **CEC_F10** | 0.002292 | 0 | 0.006114 | 0.006148 | 0.005866 | 0 | 41.94161 | 0 |
| **CEC_F11** | 0.00048 | 0.000473 | 0.000486 | 0.000486 | 0.000483 | 789.2033 | 31.14832 | 1.230663 |
| **CEC_F12** | 2.42E-06 | 2.43E-06 | 2.40E-06 | 2.41E-06 | 2.42E-06 | 149732 | 4302.336 | 0.499963 |
| **CEC_F13** | 2.35E-18 | 2.40E-18 | 2.44E-18 | 2.39E-18 | 2.37E-18 | 1.53E+17 | 33160379 | 43.51473 |
| **CEC_F14** | 0.019793 | 0.019827 | 0.019773 | 0.019808 | 0.019795 | 48.82636 | 45.96347 | 0 |

|       | ABC      | WOA      | BOA      | PSO      | EOSA     | DE       | GA       | HGSO     |
|-------|----------|----------|----------|----------|----------|----------|----------|----------|
| C_F1  | 2.73E-11 | 2.81E-11 | 2.77E-11 | 2.76E-11 | 2.76E-11 | 1.78E+09 | 6167884  | 1758730  |
| C_F2  | 2.47E-12 | 2.45E-12 | 2.46E-12 | 2.46E-12 | 2.48E-12 | 1.49E+11 | 3.96E+09 | 79257090 |
| C_F3  | 1.01E-10 | 1.01E-10 | 1.02E-10 | 1.02E-10 | 1.00E-10 | 154556.7 | 8413.72  | 511.0519 |
| C_F4  | 4.25E-06 | 4.17E-06 | 4.26E-06 | 4.27E-06 | 4.22E-06 | 30378.79 | 1195.367 | 421.332  |
| C_F5  | 0.001916 | 0.001916 | 0.001916 | 0.001916 | 0.001916 | 520      | 518.3289 | 503.618  |
| C_F6  | 0.001302 | 0.001298 | 0.001298 | 0.001298 | 0.001299 | 698.3652 | 618.1048 | 600.2262 |
| C_F7  | 0.000228 | 0.000227 | 0.000227 | 0.000226 | 0.000224 | 2044.892 | 761.7748 | 701.7483 |
| C_F8  | 0.000345 | 0.000344 | 0.000344 | 0.000343 | 0.000343 | 2147.28  | 1557.367 | 843.4343 |
| C_F9  | 0.00033  | 0.000334 | 0.000333 | 0.000332 | 0.000333 | 2255.36  | 1657.835 | 942.7569 |
| C_F10 | 2.17E-05 | 2.19E-05 | 2.19E-05 | 2.18E-05 | 2.16E-05 | 34128.95 | 22425.71 | 1997.69  |
| C_F11 | 2.19E-05 | 2.17E-05 | 2.17E-05 | 2.18E-05 | 2.17E-05 | 33964.55 | 22397.62 | 2114.091 |
| C_F12 | 0.000734 | 0.000735 | 0.000733 | 0.000732 | 0.000734 | 1228.011 | 1241.808 | 1202.002 |
| C_F13 | 0.000763 | 0.000763 | 0.000763 | 0.000763 | 0.000763 | 1305.976 | 1301.055 | 1301.188 |
| C_F14 | 0.000411 | 0.000412 | 0.000414 | 0.000413 | 0.000412 | 1775.71  | 1417.828 | 1400.543 |
| C_F15 | 2.50E-08 | 2.52E-08 | 2.49E-08 | 2.58E-08 | 2.44E-08 | 4632115  | 1589.511 | 1529.351 |
| C_F16 | 0.000604 | 0.000604 | 0.000604 | 0.000604 | 0.0006   | 1642.558 | 1642.565 | 1633.591 |
| C_F17 | 6.08E-11 | 6.27E-11 | 6.46E-11 | 6.51E-11 | 6.25E-11 | 31245659 | 972217.5 | 141272.7 |
| C_F18 | 7.19E-12 | 7.38E-12 | 7.18E-12 | 7.31E-12 | 7.33E-12 | 1.64E+10 | 14568328 | 10901519 |
| C_F19 | 1.02E-11 | 1.01E-11 | 1.04E-11 | 1.01E-11 | 1.01E-11 | 3.72E+09 | 31531.58 | 4856.103 |
| C_F20 | 6.27E-18 | 6.13E-18 | 6.15E-18 | 6.31E-18 | 6.17E-18 | 1.57E+15 | 471768.5 | 3461.329 |
| C_F21 | 1.33E-11 | 1.30E-11 | 1.34E-11 | 1.34E-11 | 1.27E-11 | 2.16E+09 | 755298.5 | 106452.7 |
| C_F22 | 8.61E-18 | 8.51E-18 | 8.22E-18 | 8.94E-18 | 8.37E-18 | 4.65E+12 | 45227.98 | 2312.382 |
| C_F23 | 4.36E-05 | 4.29E-05 | 4.27E-05 | 4.31E-05 | 4.23E-05 | 3614.32  | 2864.019 | 2427.505 |
| C_F24 | 5.36E-05 | 5.35E-05 | 5.37E-05 | 5.37E-05 | 5.32E-05 | 13879.79 | 10012.92 | 3038.901 |
| C_F25 | 0.000169 | 0.000168 | 0.000169 | 0.000169 | 0.000168 | 3201.834 | 3272.164 | 2719.248 |
| C_F26 | 7.61E-05 | 7.58E-05 | 7.66E-05 | 7.61E-05 | 7.52E-05 | 7477.168 | 4168.648 | 2831.635 |
| C_F27 | 4.39E-05 | 4.47E-05 | 4.40E-05 | 4.33E-05 | 4.34E-05 | 5538.155 | 5652.608 | 3297.419 |
| C_F28 | 2.77E-05 | 2.77E-05 | 2.73E-05 | 2.77E-05 | 2.75E-05 | 12625.29 | 20133.18 | 3338.48  |
| C_F29 | 1.01E-11 | 1.02E-11 | 1.04E-11 | 1.03E-11 | 1.02E-11 | 2.28E+10 | 19089355 | 4647940  |
| C_F30 | 1.09E-17 | 1.10E-17 | 1.06E-17 | 1.11E-17 | 1.05E-17 | 1.09E-17 | 1615135  | 49862.93 |
| AVG   | 2.07E-03 | 1.88E-03 | 2.15E-03 | 2.16E-03 | 2.15E-03 | 3.52E+15 | 1.87E+08 | 2.20E+06 |

**Table 7:** Comparison of mean values for EOSA with ABC, WOA, BOA PSO, DE, GA, and HGSO metaheuristic algorithms using the CEC functions over 20 runs and 100 population size

|        | ABC      | WOA      | BOA      | PSO      | EOSA     | DE       | GA       | HGSO     |
|--------|----------|----------|----------|----------|----------|----------|----------|----------|
| CEC_F1 | 5.33E+09 | 2.78E-11 | 2.75E-11 | 2.78E-11 | 2.75E-11 | 3.83E+09 | 70517786 | 22571997 |
| CEC_F2 | 2.06E+11 | 2.45E-12 | 2.48E-12 | 2.49E-12 | 2.48E-12 | 1.97E+11 | 5.72E+09 | 3.43E+08 |
| CEC_F3 | 265756.9 | 1.02E-10 | 1.02E-10 | 1.03E-10 | 1.01E-10 | 229059.3 | 19263.3  | 5501.943 |
| CEC_F4 | 8.86E+10 | 3.70E-12 | 3.73E-12 | 3.73E-12 | 3.71E-12 | 1.27E+11 | 1.78E+08 | 1.21E+08 |
| CEC_F5 | 20.14394 | 0.045711 | 0.045704 | 0.045704 | 0.045669 | 20.04248 | 18.59328 | 0.538232 |
| CEC_F6 | 18.67283 | 4.56E-05 | 0.005899 | 0.005873 | 0.005743 | 2.564157 | 118.74 | 0.236027 |
| CEC_F7 | 53.29714 | 0.009774 | 0.009701 | 0.009743 | 0.009735 | 54.29449 | 2.550741 | 0.1217 |
| CEC_F8 | 206645.9 | 2.40E-06 | 2.45E-06 | 2.44E-06 | 2.43E-06 | 201318.7 | 6854.159 | 555.4132 |
| CEC_F9 | 19317.58 | 2.91E-05 | 2.92E-05 | 2.94E-05 | 2.94E-05 | 17770.2 | 743.7232 | 52.63696 |
| CEC_F10 | 2.637136 | 8.98E-06 | 0.006114 | 0.006148 | 0.005868 | 0.005729 | 44.05648 | 0.000168 |
| CEC_F11 | 1068.986 | 0.000473 | 0.000486 | 0.000486 | 0.000483 | 1073.152 | 39.1105 | 3.161078 |
| CEC_F12 | 211335 | 2.43E-06 | 2.40E-06 | 2.41E-06 | 2.42E-06 | 207142.2 | 5722.981 | 404.0977 |
| CEC_F13 | 6.38E+16 | 2.40E-18 | 2.44E-18 | 2.39E-18 | 2.37E-18 | 1.53E+17 | 5.95E+13 | 4.04E+13 |
| CEC_F14 | 48.55728 | 0.019827 | 0.019773 | 0.019808 | 0.019795 | 49.02163 | 46.34748 | 1.215093 |
| C_F1 | 5.17E+09 | 2.81E-11 | 2.77E-11 | 2.76E-11 | 2.76E-11 | 3.85E+09 | 68708917 | 35387699 |
| C_F2 | 2.06E+11 | 2.45E-12 | 2.46E-12 | 2.46E-12 | 2.48E-12 | 2.07E+11 | 5.63E+09 | 4.76E+08 |
| C_F3 | 264273.1 | 1.01E-10 | 1.02E-10 | 1.02E-10 | 1.00E-10 | 229257.4 | 18720.24 | 7385.135 |
| C_F4 | 66642.05 | 4.17E-06 | 4.26E-06 | 4.27E-06 | 4.22E-06 | 60684.81 | 1629.658 | 601.2972 |
| C_F5 | 519.1411 | 0.001916 | 0.001916 | 0.001916 | 0.001916 | 520.0447 | 518.6585 | 503.8836 |
| C_F6 | 710.693 | 0.001298 | 0.001298 | 0.001298 | 0.001299 | 713.7333 | 620.3631 | 600.7176 |
| C_F7 | 2558.992 | 0.000227 | 0.000227 | 0.000226 | 0.000224 | 2521.263 | 776.6596 | 707.6167 |
| C_F8 | 1817.585 | 0.000344 | 0.000344 | 0.000343 | 0.000343 | 2229.991 | 1581.796 | 858.1452 |
| C_F9 | 1918.445 | 0.000334 | 0.000333 | 0.000332 | 0.000333 | 2357.754 | 1683.917 | 954.2328 |
| C_F10 | 23020 | 2.19E-05 | 2.19E-05 | 2.18E-05 | 2.16E-05 | 34606.45 | 23387.87 | 2446.736 |
| C_F11 | 22642.04 | 2.17E-05 | 2.17E-05 | 2.18E-05 | 2.17E-05 | 34535.5 | 23444.82 | 2538.994 |
| C_F12 | 1200.547 | 0.000735 | 0.000733 | 0.000732 | 0.000734 | 1231.839 | 1244.155 | 1203.938 |
| C_F13 | 1304.738 | 0.000763 | 0.000763 | 0.000763 | 0.000763 | 1307.094 | 1301.152 | 1301.302 |
| C_F14 | 1915.752 | 0.000412 | 0.000414 | 0.000413 | 0.000412 | 1909.953 | 1422.289 | 1401.656 |
| C_F15 | 3991227 | 2.52E-08 | 2.49E-08 | 2.58E-08 | 2.44E-08 | 8250976 | 5798.619 | 2590.23 |
| C_F16 | 1631.447 | 0.000604 | 0.000604 | 0.000604 | 0.0006 | 1642.966 | 1642.567 | 1635.098 |
| C_F17 | 6.52E+08 | 6.27E-11 | 6.46E-11 | 6.51E-11 | 6.25E-11 | 2.35E+08 | 11945024 | 6293094 |
| C_F18 | 3.41E+10 | 7.38E-12 | 7.18E-12 | 7.31E-12 | 7.34E-12 | 3.17E+10 | 3.32E+08 | 1.25E+08 |
| C_F19 | 1.01E+10 | 1.01E-11 | 1.04E-11 | 1.01E-11 | 1.01E-11 | 1.07E+10 | 26262424 | 14398954 |
| C_F20 | 3.09E+15 | 6.13E-18 | 6.15E-18 | 6.31E-18 | 6.17E-18 | 9.07E+15 | 3.96E+12 | 8.23E+12 |
| C_F21 | 6.69E+09 | 1.30E-11 | 1.34E-11 | 1.34E-11 | 1.27E-11 | 5.96E+09 | 29193444 | 17906516 |
| C_F22 | 5.71E+14 | 8.51E-18 | 8.22E-18 | 8.94E-18 | 8.37E-18 | 7.38E+14 | 7.24E+11 | 5.42E+12 |
| C_F23 | 5648.371 | 4.29E-05 | 4.27E-05 | 4.31E-05 | 4.23E-05 | 4936.6 | 2915.359 | 2442.92 |
| C_F24 | 10545.33 | 5.35E-05 | 5.37E-05 | 5.37E-05 | 5.32E-05 | 14160.14 | 10305.09 | 3177.754 |
| C_F25 | 3470.654 | 0.000168 | 0.000169 | 0.000169 | 0.000168 | 3362.374 | 3294.278 | 2742.184 |
| C_F26 | 8109.353 | 7.58E-05 | 7.66E-05 | 7.61E-05 | 7.52E-05 | 8290.179 | 4250.187 | 2857.924 |
| C_F27 | 7333.635 | 4.47E-05 | 4.40E-05 | 4.33E-05 | 4.34E-05 | 6777.938 | 5851.102 | 3486.684 |
| C_F28 | 17319.94 | 2.77E-05 | 2.73E-05 | 2.77E-05 | 2.75E-05 | 14334 | 20220.09 | 3719.304 |
| C_F29 | 2.68E+10 | 1.02E-11 | 1.04E-11 | 1.03E-11 | 1.02E-11 | 3.07E+10 | 2.13E+08 | 93832603 |
| C_F30 | 3.29E+14 | 1.10E-17 | 1.06E-17 | 1.11E-17 | 1.05E-17 | 1.09E-17 | 6.25E+11 | 8.25E+11 |
| AVG | 1.54E+15 | 1.89E-03 | 2.15E-03 | 2.16E-03 | 2.15E-03 | 3.71E+15 | 1.47E+12 | 1.25E+12 |

**Table 8:** Comparison of standard deviation values for EOSA with ABC, WOA, BOA PSO, DE, GA, and HGSO metaheuristic algorithms using the CEC functions over 20 runs and 100 population size

|  | ABC | WOA | BOA | PSO | EOSA | DE | GA | HGSO |
|---|---|---|---|---|---|---|---|---|
| **CEC_F1** | 8.44E+08 | 3.39E-27 | 4.36E-27 | 3.88E-27 | 1.46E-15 | 2.13E+09 | 3.1E+08 | 2.43E+08 |
| **CEC_F2** | 1.3E+10 | 4.64E-28 | 3.03E-28 | 2.83E-28 | 9.11E-17 | 3.92E+10 | 7.53E+09 | 4.41E+09 |
| **CEC_F3** | 124317.8 | 1.36E-26 | 2.07E-26 | 2.13E-26 | 3.19E-14 | 80504.07 | 23773.8 | 94634.65 |
| **CEC_F4** | 9.64E+09 | 5.65E-28 | 6.26E-28 | 7.88E-28 | 1.35E-16 | 7.2E+09 | 2.26E+09 | 1.59E+09 |
| **CEC_F5** | 0.957905 | 5.90E-18 | 4.16E-18 | 5.90E-18 | 9.25E-07 | 0.146082 | 0.531952 | 2.67635 |
| **CEC_F6** | 28.93125 | 0.000486 | 1.08E-18 | 6.94E-19 | 8.98E-06 | 9.64147 | 2.394845 | 2.535846 |
| **CEC_F7** | 3.287874 | 9.54E-19 | 1.04E-18 | 1.13E-18 | 5.86E-07 | 9.165075 | 1.842338 | 1.222283 |
| **CEC_F8** | 13463.5 | 3.18E-22 | 3.81E-22 | 3.18E-22 | 7.16E-11 | 44270.92 | 7394.998 | 5414.126 |
| **CEC_F9** | 1125.511 | 4.91E-21 | 4.91E-21 | 4.57E-21 | 7.61E-09 | 2475.684 | 812.3379 | 525.8539 |
| **CEC_F10** | 8.081229 | 0.000189 | 9.54E-19 | 1.04E-18 | 1.80E-05 | 0.079924 | 3.439061 | 0.003118 |
| **CEC_F11** | 67.00617 | 7.59E-20 | 6.78E-20 | 8.13E-20 | 2.51E-08 | 203.6223 | 38.53338 | 22.77025 |
| **CEC_F12** | 13017.52 | 2.75E-22 | 2.12E-22 | 3.18E-22 | 7.17E-11 | 40966.34 | 7434.063 | 4920.48 |
| **CEC_F13** | 1.56E+16 | 3.08E-34 | 3.08E-34 | 3.27E-34 | 6.53E-22 | 16 | 1.16E+15 | 7.79E+14 |
| **CEC_F14** | 2.18245 | 2.26E-18 | 2.60E-18 | 2.26E-18 | 1.36E-06 | 0.160205 | 0.52921 | 7.091001 |
| **C_F1** | 9.38E+08 | 5.65E-27 | 4.04E-27 | 3.72E-27 | 1.54E-15 | 2.18E+09 | 3.19E+08 | 3.53E+08 |
| **C_F2** | 1.27E+10 | 3.84E-28 | 3.43E-28 | 2.63E-28 | 5.13E-17 | 3.91E+10 | 7.35E+09 | 5.18E+09 |
| **C_F3** | 97428.44 | 1.42E-26 | 1.42E-26 | 1.42E-26 | 3.41E-14 | 85524.47 | 21249.39 | 92545.77 |
| **C_F4** | 7311.002 | 6.56E-22 | 5.08E-22 | 7.41E-22 | 1.73E-09 | 25164.55 | 2379.42 | 1832.528 |
| **C_F5** | 23.24212 | 3.79E-19 | 2.93E-19 | 2.49E-19 | 7.25E-10 | 0.144908 | 0.504494 | 1.673248 |
| **C_F6** | 31.98589 | 1.52E-19 | 1.30E-19 | 1.63E-19 | 8.59E-09 | 10.84148 | 6.159877 | 4.5231 |
| **C_F7** | 141.8918 | 4.07E-20 | 3.39E-20 | 3.25E-20 | 7.57E-09 | 329.3684 | 66.00672 | 52.04919 |
| **C_F8** | 156.4804 | 3.79E-20 | 3.52E-20 | 3.52E-20 | 4.04E-08 | 78.9143 | 44.58426 | 95.90718 |
| **C_F9** | 158.7789 | 4.34E-20 | 3.25E-20 | 4.88E-20 | 1.49E-08 | 82.38662 | 45.19802 | 86.11494 |
| **C_F10** | 3268.026 | 2.20E-21 | 2.54E-21 | 1.86E-21 | 2.54E-09 | 423.4968 | 1633.723 | 2997.41 |
| **C_F11** | 3252.009 | 4.24E-21 | 3.05E-21 | 3.90E-21 | 1.13E-09 | 560.3687 | 1641.823 | 2811.823 |
| **C_F12** | 54.43402 | 1.03E-19 | 9.76E-20 | 1.03E-19 | 6.74E-08 | 4.051227 | 3.398669 | 5.949288 |
| **C_F13** | 58.4083 | 8.67E-20 | 1.03E-19 | 1.14E-19 | 9.74E-10 | 0.786165 | 0.315701 | 0.26207 |
| **C_F14** | 88.99855 | 7.05E-20 | 4.61E-20 | 6.51E-20 | 2.92E-09 | 91.06467 | 19.00835 | 10.87745 |
| **C_F15** | 989141.6 | 4.63E-24 | 2.98E-24 | 4.14E-24 | 3.48E-11 | 2442085 | 78999.93 | 20600.54 |
| **C_F16** | 73.05729 | 8.13E-20 | 4.34E-20 | 8.67E-20 | 2.35E-07 | 0.325375 | 0.037071 | 1.887207 |
| **C_F17** | 2E+08 | 9.37E-27 | 1.62E-26 | 9.69E-27 | 8.98E-15 | 3.7E+08 | 62967751 | 65905146 |
| **C_F18** | 4.48E+09 | 1.62E-27 | 1.29E-27 | 1.21E-27 | 1.91E-15 | 1.26E+10 | 1.79E+09 | 1.26E+09 |
| **C_F19** | 2.5E+09 | 1.13E-27 | 1.37E-27 | 1.37E-27 | 1.68E-15 | 6.61E+09 | 3.95E+08 | 2.2E+08 |
| **C_F20** | 2.2E+15 | 1.00E-33 | 1.16E-33 | 9.63E-34 | 2.07E-21 | 7.23E+15 | 8.15E+13 | 1.33E+14 |
| **C_F21** | 1.71E+09 | 1.94E-27 | 2.58E-27 | 2.02E-27 | 1.70E-15 | 4.08E+09 | 2.93E+08 | 2.81E+08 |
| **C_F22** | 6.8E+14 | 1.16E-33 | 1.39E-33 | 8.47E-34 | 2.69E-21 | 1.44E+15 | 1.49E+13 | 1.03E+14 |
| **C_F23** | 505.2743 | 6.10E-21 | 5.76E-21 | 6.10E-21 | 5.58E-09 | 1319.089 | 159.8902 | 144.9136 |
| **C_F24** | 1142.609 | 8.13E-21 | 8.13E-21 | 6.10E-21 | 7.63E-09 | 294.4141 | 523.598 | 873.6135 |
| C_F25 | 172.4484 | 2.17E-20 | 2.57E-20 | 2.71E-20 | 1.41E-08 | 175.9663 | 26.39227 | 92.35979 |
| C_F26 | 447.5749 | 8.81E-21 | 7.45E-21 | 8.81E-21 | 1.23E-08 | 609.1627 | 214.4841 | 201.5054 |
| C_F27 | 555.7132 | 6.44E-21 | 5.08E-21 | 4.74E-21 | 9.68E-09 | 1223.351 | 231.5779 | 574.4783 |
| C_F28 | 1546.707 | 3.73E-21 | 5.76E-21 | 4.07E-21 | 2.74E-09 | 1742.996 | 532.2733 | 2264.442 |
| C_F29 | 2.8E+09 | 1.86E-27 | 1.62E-27 | 1.37E-27 | 1.54E-15 | 6.16E+09 | 1.15E+09 | 9.03E+08 |
| C_F30 | 4.95E+14 | 1.54E-33 | 1.62E-33 | 1.39E-33 | 3.53E-21 | 1.05E-21 | 1.24E+13 | 1.7E+13 |
| AVG | 4.31E+14 | 1.54E-05 | 2.45E-19 | 2.74E-19 | 6.90E-07 | 1.97E+14 | 2.89E+13 | 2.34E+13 |

**Table 9:** Comparison of worst values for EOSA with ABC, WOA, BOA PSO, DE, GA, and HGSO metaheuristic algorithms using the CEC functions over 20 runs and 100 population size

| | ABC | WOA | BOA | PSO | EOSA3 | DE | GA | HGSO |
|---|---|---|---|---|---|---|---|---|
| CEC_F1 | 1.12E+10 | 2.78E-11 | 2.75E-11 | 2.78E-11 | 2.75E-11 | 1.04E+10 | 5.59E+09 | 4.16E+09 |
| CEC_F2 | 2.59E+11 | 2.45E-12 | 2.48E-12 | 2.49E-12 | 2.48E-12 | 2.56E+11 | 1.38E+11 | 8.22E+10 |
| CEC_F3 | 2687918 | 1.02E-10 | 1.02E-10 | 1.03E-10 | 1.02E-10 | 758263.8 | 210260.8 | 1949480 |
| CEC_F4 | 1.36E+11 | 3.70E-12 | 3.73E-12 | 3.73E-12 | 3.71E-12 | 1.33E+11 | 4.71E+10 | 2.92E+10 |
| CEC_F5 | 21.52029 | 0.045711 | 0.045704 | 0.045704 | 0.04568 | 21.17672 | 21.51461 | 20.26211 |
| CEC_F6 | 131.4872 | 0.005909 | 0.005899 | 0.005873 | 0.005825 | 90.86362 | 129.6001 | 44.31723 |
| CEC_F7 | 67.33259 | 0.009774 | 0.009701 | 0.009743 | 0.009748 | 66.20478 | 35.20829 | 21.53974 |
| CEC_F8 | 262494.3 | 2.40E-06 | 2.45E-06 | 2.44E-06 | 2.43E-06 | 262963.7 | 136488.5 | 89507.12 |
| CEC_F9 | 23595.96 | 2.91E-05 | 2.92E-05 | 2.94E-05 | 2.95E-05 | 23589.75 | 14244.41 | 8958.105 |
| CEC_F10 | 63.28519 | 0.004209 | 0.006114 | 0.006148 | 0.006093 | 1.718978 | 64.79516 | 0.06861 |
| CEC_F11 | 1327.766 | 0.000473 | 0.000486 | 0.000486 | 0.000484 | 1347.94 | 717.9246 | 422.4248 |
| CEC_F12 | 262772.6 | 2.43E-06 | 2.40E-06 | 2.41E-06 | 2.42E-06 | 262112.5 | 134429.3 | 88298.64 |
| CEC_F13 | 1.54E+17 | 2.40E-18 | 2.44E-18 | 2.39E-18 | 2.38E-18 | 1.53E+17 | 2.58E+16 | 1.71E+16 |
| CEC_F14 | 49.52848 | 0.019827 | 0.019773 | 0.019808 | 0.019824 | 49.48733 | 49.33462 | 49.38548 |
| C_F1 | 1.15E+10 | 2.81E-11 | 2.77E-11 | 2.76E-11 | 2.77E-11 | 1.11E+10 | 5.67E+09 | 6.19E+09 |
| C_F2 | 2.58E+11 | 2.45E-12 | 2.46E-12 | 2.46E-12 | 2.48E-12 | 2.59E+11 | 1.33E+11 | 1.01E+11 |
| C_F3 | 2103257 | 1.01E-10 | 1.02E-10 | 1.02E-10 | 1.01E-10 | 917649.9 | 250629.4 | 1908977 |
| C_F4 | 105914.3 | 4.17E-06 | 4.26E-06 | 4.27E-06 | 4.26E-06 | 105781.7 | 46273.45 | 30893.76 |
| C_F5 | 521.5171 | 0.001916 | 0.001916 | 0.001916 | 0.001916 | 521.1812 | 521.5299 | 519.7741 |
| C_F6 | 730.2922 | 0.001298 | 0.001298 | 0.001298 | 0.001299 | 730.5756 | 696.6942 | 670.6866 |
| C_F7 | 3079.45 | 0.000227 | 0.000227 | 0.000226 | 0.000224 | 3029.567 | 1923.113 | 1559.444 |
| C_F8 | 2406.95 | 0.000344 | 0.000344 | 0.000343 | 0.000343 | 2400.625 | 2074.284 | 1962.092 |
| C_F9 | 2506.628 | 0.000334 | 0.000333 | 0.000332 | 0.000334 | 2506.397 | 2177.013 | 2044.832 |
| C_F10 | 36158.47 | 2.19E-05 | 2.19E-05 | 2.18E-05 | 2.17E-05 | 36205.08 | 34710.49 | 33711.3 |
| C_F11 | 36240.38 | 2.17E-05 | 2.17E-05 | 2.18E-05 | 2.17E-05 | 36030.53 | 34829.66 | 33964.02 |
| C_F12 | 1264.28 | 0.000735 | 0.000733 | 0.000732 | 0.000734 | 1254.941 | 1261.787 | 1254.078 |
| C_F13 | 1308.329 | 0.000763 | 0.000763 | 0.000763 | 0.000763 | 1308.369 | 1306.122 | 1304.812 |
| C_F14 | 2070.763 | 0.000412 | 0.000414 | 0.000413 | 0.000412 | 2042.449 | 1748.141 | 1589.229 |
| C_F15 | 10490478 | 2.52E-08 | 2.49E-08 | 2.58E-08 | 2.51E-08 | 10914694 | 1748135 | 444906.5 |
| C_F16 | 1643.261 | 0.000604 | 0.000604 | 0.000604 | 0.000604 | 1643.697 | 1643.287 | 1643.891 |
| C_F17 | 2.25E+09 | 6.27E-11 | 6.46E-11 | 6.51E-11 | 6.27E-11 | 2.25E+09 | 1.17E+09 | 1.18E+09 |
| C_F18 | 5.88E+10 | 7.38E-12 | 7.18E-12 | 7.31E-12 | 7.35E-12 | 6.01E+10 | 3.2E+10 | 2.18E+10 |
| C_F19 | 2.73E+10 | 1.01E-11 | 1.04E-11 | 1.01E-11 | 1.01E-11 | 2.44E+10 | 8.51E+09 | 4.48E+09 |
| C_F20 | 2.05E+16 | 6.13E-18 | 6.15E-18 | 6.31E-18 | 6.21E-18 | 2.19E+16 | 1.81E+15 | 2.51E+15 |
| C_F21 | 1.81E+10 | 1.30E-11 | 1.34E-11 | 1.34E-11 | 1.27E-11 | 1.79E+10 | 6.09E+09 | 5.93E+09 |
| C_F22 | 7.01E+15 | 8.51E-18 | 8.22E-18 | 8.94E-18 | 8.43E-18 | 6.17E+15 | 3.32E+14 | 2.19E+15 |
| C_F23 | 8511.456 | 4.29E-05 | 4.27E-05 | 4.31E-05 | 4.24E-05 | 8622.02 | 5437.081 | 4831.978 |
| C_F24 | 15051.17 | 5.35E-05 | 5.37E-05 | 5.37E-05 | 5.34E-05 | 15014.57 | 14393.9 | 13420.02 |
| C_F25 | 3955.273 | 0.000168 | 0.000169 | 0.000169 | 0.000168 | 3923.978 | 3536.828 | 3525.063 |
| C_F26 | 9478.512 | 7.58E-05 | 7.66E-05 | 7.61E-05 | 7.55E-05 | 9528.773 | 7271.734 | 6069.047 |
| C_F27 | 10435.93 | 4.47E-05 | 4.40E-05 | 4.33E-05 | 4.35E-05 | 10466.53 | 7279.321 | 8191.818 |
| C_F28 | 23982.63 | 2.77E-05 | 2.73E-05 | 2.77E-05 | 2.76E-05 | 22936.38 | 24198.69 | 23592.56 |
| C_F29 | 4.24E+10 | 1.02E-11 | 1.04E-11 | 1.03E-11 | 1.02E-11 | 4.14E+10 | 2.11E+10 | 1.53E+10 |
| C_F30 | 5.34E+15 | 1.10E-17 | 1.06E-17 | 1.11E-17 | 1.06E-17 | 1.09E-17 | 2.77E+14 | 3.76E+14 |
| AVG | 4.25E+15 | 2.11E-03 | 2.15E-03 | 2.16E-03 | 2.15E-03 | 4.12E+15 | 6.41E+14 | 5.04E+14 |

**Table 10:** Comparison of median values for EOSA with ABC, WOA, BOA PSO, DE, GA, and HGSO metaheuristic algorithms using the CEC functions over 20 runs and 100 population size

| | ABC | WOA | BOA | PSO | EOSA | DE | GA | HGSO |
|---|---|---|---|---|---|---|---|---|
| CEC_F1 | 4.97E+09 | 2.78E-11 | 2.75E-11 | 2.78E-11 | 2.75E-11 | 3.16E+09 | 17405089 | 7.98E-53 |
| CEC_F2 | 2.02E+11 | 2.45E-12 | 2.48E-12 | 2.49E-12 | 2.48E-12 | 1.95E+11 | 4.44E+09 | 1.11E-54 |
| CEC_F3 | 251561 | 1.02E-10 | 1.02E-10 | 1.03E-10 | 1.01E-10 | 211966.5 | 12804.61 | 3.54E-59 |
| CEC_F4 | 8.5E+10 | 3.70E-12 | 3.73E-12 | 3.73E-12 | 3.71E-12 | 1.3E+11 | 5359283 | 98.87237 |
| CEC_F5 | 20.02451 | 0.045711 | 0.045704 | 0.045704 | 0.045669 | 20.00025 | 18.40464 | 6.22E-16 |
| CEC_F6 | 4.21278 | 0 | 0.005899 | 0.005873 | 0.005742 | 0.000395 | 118.3358 | 0 |
| CEC_F7 | 52.42071 | 0.009774 | 0.009701 | 0.009743 | 0.009735 | 54.26327 | 2.251702 | 0 |
| CEC_F8 | 202970.3 | 2.40E-06 | 2.45E-06 | 2.44E-06 | 2.43E-06 | 193545.9 | 5656.178 | 0 |
| CEC_F9 | 19028.6 | 2.91E-05 | 2.92E-05 | 2.94E-05 | 2.94E-05 | 17241.84 | 608.2115 | 0.001273 |
| CEC_F10 | 0.15593 | 0 | 0.006114 | 0.006148 | 0.005866 | 3.57E-06 | 42.45588 | 0 |
| CEC_F11 | 1050.116 | 0.000473 | 0.000486 | 0.000486 | 0.000483 | 1037.34 | 32.65625 | 1.305509 |
| CEC_F12 | 208014.3 | 2.43E-06 | 2.40E-06 | 2.41E-06 | 2.42E-06 | 203259.8 | 4502.993 | 0.499963 |
| CEC_F13 | 5.70E+16 | 2.40E-18 | 2.44E-18 | 2.39E-18 | 2.37E-18 | 1.53E+17 | 1.44E+09 | 43.7952 |
| CEC_F14 | 48.56286 | 0.019827 | 0.019773 | 0.019808 | 0.019795 | 48.98768 | 46.16493 | 0 |
| C_F1 | 4.81E+09 | 2.81E-11 | 2.77E-11 | 2.76E-11 | 2.76E-11 | 3.07E+09 | 15800257 | 2539288 |
| C_F2 | 2.02E+11 | 2.45E-12 | 2.46E-12 | 2.46E-12 | 2.48E-12 | 2.07E+11 | 4.48E+09 | 83310110 |
| C_F3 | 251898.3 | 1.01E-10 | 1.02E-10 | 1.02E-10 | 1.00E-10 | 207464.7 | 12286.88 | 569.2232 |
| C_F4 | 63680.08 | 4.17E-06 | 4.26E-06 | 4.27E-06 | 4.22E-06 | 53106.6 | 1289.325 | 421.5314 |
| C_F5 | 520.0264 | 0.001916 | 0.001916 | 0.001916 | 0.001916 | 520 | 518.4432 | 503.6196 |
| C_F6 | 710.5576 | 0.001298 | 0.001298 | 0.001298 | 0.001299 | 713.8983 | 619.0061 | 600.2345 |
| C_F7 | 2526.215 | 0.000227 | 0.000227 | 0.000226 | 0.000224 | 2476.466 | 766.6583 | 701.7771 |
| C_F8 | 1756.355 | 0.000344 | 0.000344 | 0.000343 | 0.000343 | 2205.942 | 1567.959 | 844.6982 |
| C_F9 | 1857.348 | 0.000334 | 0.000333 | 0.000332 | 0.000333 | 2351.115 | 1671.562 | 944.6468 |
| C_F10 | 21565.89 | 2.19E-05 | 2.19E-05 | 2.18E-05 | 2.16E-05 | 34546.78 | 22826.18 | 2051.625 |
| C_F11 | 21202.09 | 2.17E-05 | 2.17E-05 | 2.18E-05 | 2.17E-05 | 34411.18 | 23005.81 | 2162.177 |
| C_F12 | 1200.098 | 0.000735 | 0.000733 | 0.000732 | 0.000734 | 1230.866 | 1243.101 | 1202.801 |
| C_F13 | 1307.277 | 0.000763 | 0.000763 | 0.000763 | 0.000763 | 1306.991 | 1301.12 | 1301.257 |
| C_F14 | 1908.945 | 0.000412 | 0.000414 | 0.000413 | 0.000412 | 1894.814 | 1419.233 | 1400.575 |
| C_F15 | 3574560 | 2.52E-08 | 2.49E-08 | 2.58E-08 | 2.44E-08 | 8643243 | 1592.3 | 1529.579 |
| C_F16 | 1633.813 | 0.000604 | 0.000604 | 0.000604 | 0.0006 | 1643.018 | 1642.565 | 1634.636 |
| C_F17 | 5.76E+08 | 6.27E-11 | 6.46E-11 | 6.51E-11 | 6.25E-11 | 92363316 | 2870311 | 249692.2 |
| C_F18 | 3.25E+10 | 7.38E-12 | 7.18E-12 | 7.31E-12 | 7.33E-12 | 2.81E+10 | 50256334 | 14649820 |
| C_F19 | 9.14E+09 | 1.01E-11 | 1.04E-11 | 1.01E-11 | 1.01E-11 | 8.3E+09 | 218340.8 | 6335.723 |
| C_F20 | 2.34E+15 | 6.13E-18 | 6.15E-18 | 6.31E-18 | 6.17E-18 | 7.03E+15 | 11716450 | 3736.184 |
| C_F21 | 6.05E+09 | 1.30E-11 | 1.34E-11 | 1.34E-11 | 1.27E-11 | 4.54E+09 | 2438312 | 166857.2 |
| C_F22 | 3.46E+14 | 8.51E-18 | 8.22E-18 | 8.94E-18 | 8.37E-18 | 8.57E+13 | 1945742 | 2390.839 |
| C_F23 | 5473.74 | 4.29E-05 | 4.27E-05 | 4.31E-05 | 4.23E-05 | 4454.57 | 2878.711 | 2428.272 |
| C_F24 | 10065.35 | 5.35E-05 | 5.37E-05 | 5.37E-05 | 5.32E-05 | 14042.95 | 10141.77 | 3059.551 |
| C_F25 | 3444.507 | 0.000168 | 0.000169 | 0.000169 | 0.000168 | 3299.447 | 3287.133 | 2726.605 |
| C_F26 | 8002.961 | 7.58E-05 | 7.66E-05 | 7.61E-05 | 7.52E-05 | 8208.712 | 4185.872 | 2835.785 |
| C_F27 | 7171.65 | 4.47E-05 | 4.40E-05 | 4.33E-05 | 4.34E-05 | 6358.864 | 5809.833 | 3329.576 |
| C_F28 | 16715.8 | 2.77E-05 | 2.73E-05 | 2.77E-05 | 2.75E-05 | 13773.55 | 20133.18 | 3352.717 |
| C_F29 | 2.57E+10 | 1.02E-11 | 1.04E-11 | 1.03E-11 | 1.02E-11 | 3E+10 | 47619482 | 5734789 |
| C_F30 | 1.85E+14 | 1.10E-17 | 1.06E-17 | 1.11E-17 | 1.05E-17 | 1.09E-17 | 25582427 | 76134.32 |
| AVG | 1.36E+15 | 1.88E-03 | 2.15E-03 | 2.16E-03 | 2.15E-03 | 3.65E+15 | 2.40E+08 | 2.43E+06 |

Similarly, the result for the worst values for the 44 functions, as shown in its corresponding table, also reveals an interesting performance for the proposed EOSA. We discovered that EOSA obtained the same values with WOA, BOA, and PSO using C5, C13, and C16 functions, whereas the corresponding values for ABC, DE, GA, and HGSO were significantly large. Also, EOSA leaped in-between in its values when compared with WOA, BOA, and PSO based on CEC_F1-4, CEC_F7-8, CEC_F10-12, CEC_F14, C1, C3, C8-9, C11-12, C14-15, C17-20, C22, C25, and C29-30functions, whereas it records an overall good performance for CEC05-06, CEC13, C4, C7, C10, C21, C23-24, and C26-28.

Mean values for EOSA, ABC, WOA, BOA, and PSO demonstrated a very close outcome with no clear leading algorithm in the cases of CEC01-04, CEC06, CEC08-10, CEC12, CEC14, C1, C4, C8-9, C11-12, C14, C18-20, C22, C25, and C27-29functions. But for the CEC05, CEC07, CEC11, CEC13, C3, C7, C10, C15-17, C21, C23-24, C26, and C30 functions, the mean values of EOSA outperformed all related algorithms including ABC, DE, GA, and HGSO, while it lags behind for the C2 and C6 functions. We also observed that the same values were obtained for EOSA, WOA, BOA, and PSO for the C5 and C13.

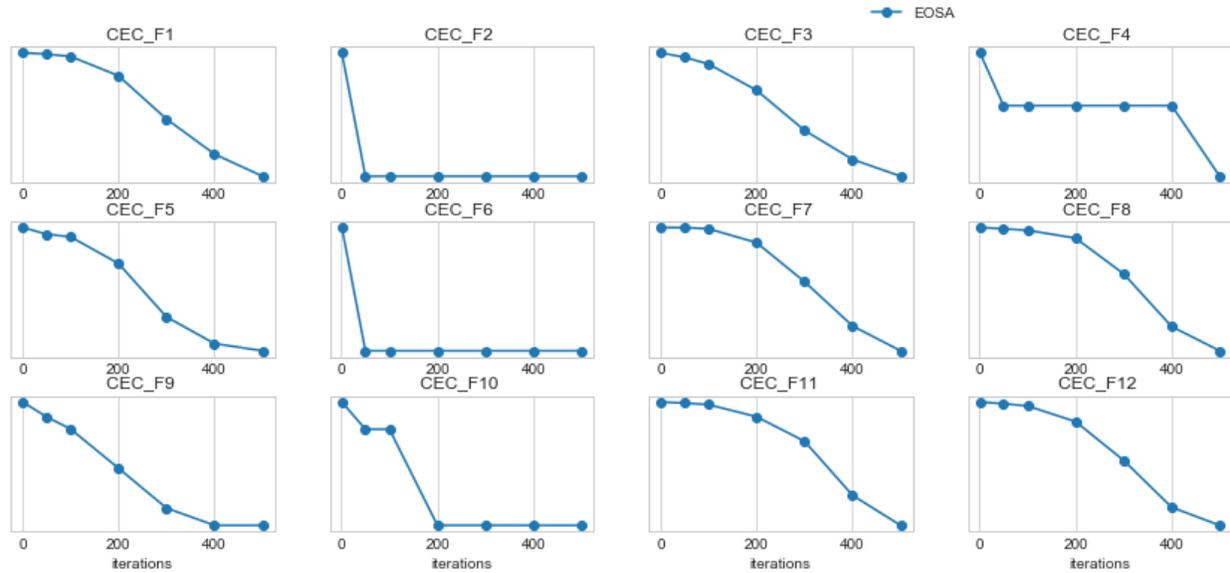

**Figure 8:** Convergent curves of EOSA on some selected CEC benchmark functions over 1, 50, 100, 200, 300, 400 and 500 epochs

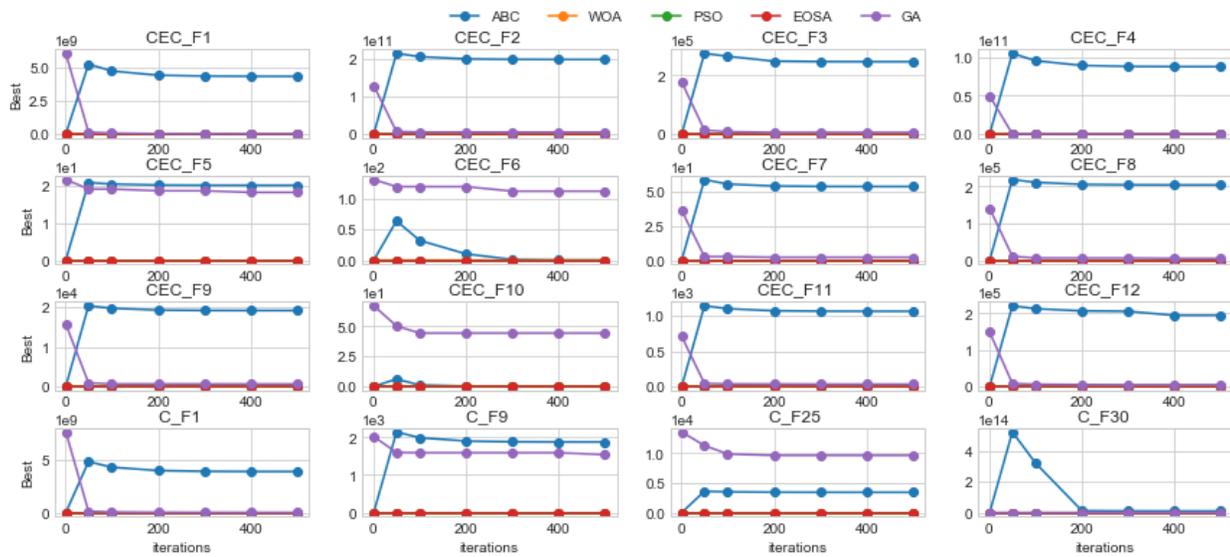

**Figure 9:** Convergent curves of EOSA and related optimization algorithms on some selected CEC functions

In Figure 8, the curves of the convergence of EOSA on the solutions of CEC benchmark functions are shown. We graphed the optimized solutions over 1, 50, 100, 200, 300, 400, and 500 epochs. To demonstrate this, we plot the curves of CEC_F1-14 and C_F1. The outcome of the pattern of the curves showed that the solutions were well optimized over the epochs. The convergence of the EOSA algorithm based on the best fit as compared with the best fits of ABC, WOA, BOA, PSO, DE, GA, and HGSO we plotted as shown in Figure 9. All the curves representing each optimisation algorithm showed a descent from high to low based on their best values. Although the curves of ABC and occasionally that of GA were often seen overshooting in values compared with others, we confirm that this is not unrelated to the significantly large values obtained by these algorithms (ABC and GA, with DE and HGSO inclusive). The curves of EOSA, WOA, BOA, and PSO appear to lie low, though with marginal descent overshadowed by ABC and GA continuously.

## 5.4 Computational requirement for EOSA and related algorithms on benchmark and CEC functions

The computational time required to run the optimization algorithms discussed in previous subsections was also recorded and reported in this subsection. We took an average of the computation time for all the forty-seven (47) standard benchmark functions and the forty-four (44) CEC-based functions. The outcome of these averages for ABC, WOA, BOA, PSO, DE, GA, and HGSO compared with EOSA are listed in Table 11. We discovered that the computational requirement of EOSA reports a minimal CPU time compared with other algorithms.

**Table 11:** Average computational requirements of ABC, WOA, BOA PSO, QSO, DE, GA, and HGSO for forty-seven (47) benchmarks and forty-four (44) CEC for 500 runs and 100 population size

|  | ABC | WOA | BOA | PSO | EOSA | DE | GA | HGSO |
|---|---|---|---|---|---|---|---|---|
| **Benchmark functions** | 450.9394 | 357.6453 | 418.0029 | 353.3408 | 132.1047 | 441.237 | 365.8788 | 442.8186 |
| **CEC** | 450.9394 | 357.6453 | 418.0029 | 353.3408 | 132.1047 | 441.237 | 365.8788 | 442.8186 |

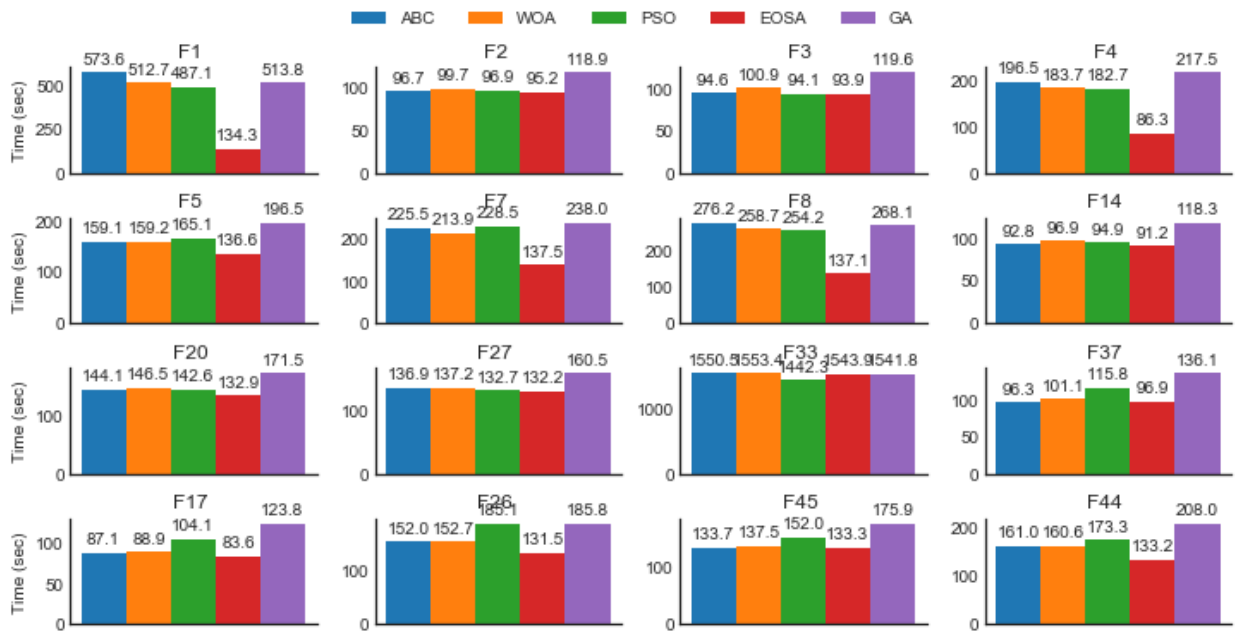

**Figure 10:** A graphical illustration of computational time required for the execution of EOSA compared with ABC, WOA, PSO, and GA for the standard benchmark functions

The computational requirement for executing all the algorithms on the standard benchmark functions was graphed as shown in Figure 10. We randomly selected some of these functions: F1, F2, F3, F4, F5, F7, F8, F14, F20, F27, F33, F37, F17, F26, D45, and F44. Although EOSA consumed the lowest CPU time in all cases, we noticed that the discrepancies in F27, F33, F2, and F14 were quite marginal. On the other hand, EOSA's computational requirements in F1, F4, and F8 were significantly low compared with related optimization algorithms.

Similarly, In Figure 11, the computational requirement for CEC-based functions was illustrated for some selected functions namely CEC01, CEC02, CEC03, CEC04, CEC05, CEC06, CEC07, CEC10, CEC11, CEC12, C1, C9, C25, and C30. In all cases, the CPU time for training EOSA showed to be lower than other related algorithms. While EOSA showed required less CPU time, GA and ABC were more demanding for this same computational resource. We note that the unusual computational time accounted for in some of the algorithms might not be unconnected with occasions when several algorithms experimented on the same system.

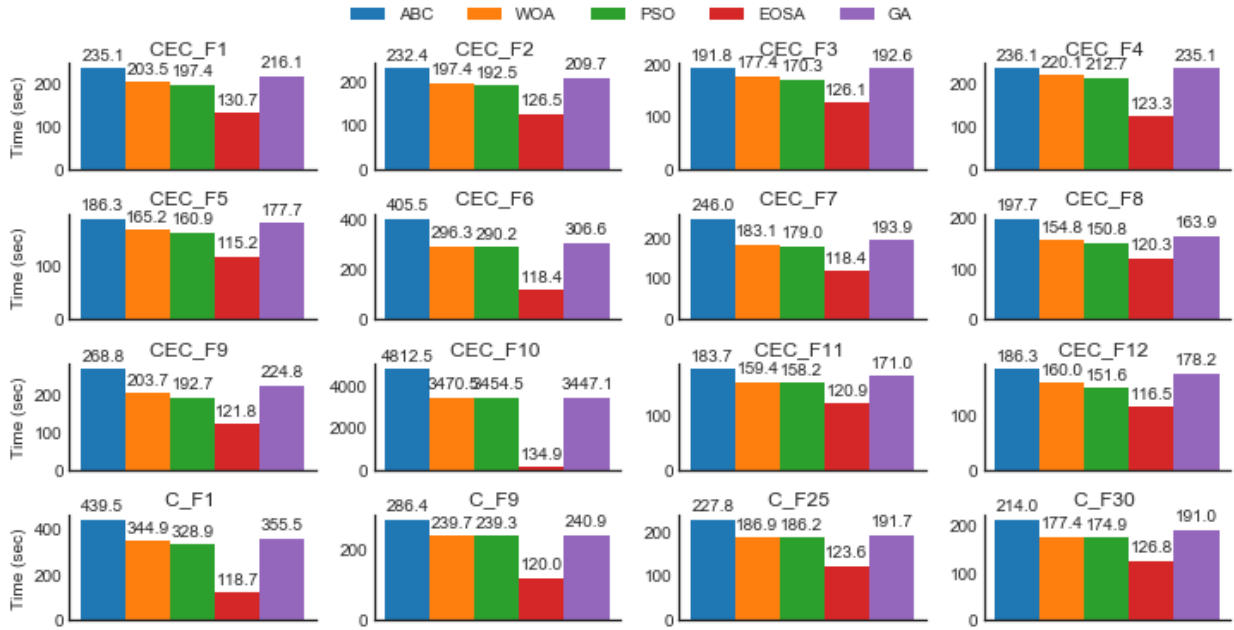

**Figure 11:** A graphical illustration of computational time required for the execution of EOSA compared with ABC, WOA, PSO, and GA for the CEC functions

### 5.5 Comparing the Performance of EOSA with Similar Methods Using Statistical Test

Using the results obtained from the forty-seven (47) standard benchmark functions, we validated the performance gain of EOSA as against those of ABC, WOA, BOA, PSO, DE, GA, and HGSO using statistical analysis. To achieve this, the Friedman mean the rank test was carried out, and the result obtained is as shown in Table 12. The results showed that the proposed EOSA method ranks best above all other methods by yielding the mean rank of 1.60. The PSO, WOA, and BOA trails after HGSO, GA, ABC, and DE follow in that order.

**Table 12:** Friedman mean Ranks test for EOSA compared with similar optimization algorithms

| Algorithm | Mean rank | General rank |
|---|---|---|
| ABC | 7.10 | 6 |
| WOA | 2.90 | 3 |
| BOA | 2.90 | 3 |
| PSO | 2.79 | 2 |
| EOSA | 1.60 | 1 |
| DE | 7.57 | 6 |
| GA | 6.12 | 5 |
| HGSO | 5.02 | 4 |

The test statistics ($\chi^2$) result for the Friedman test revealed that there was an overall statistically significant difference between the mean ranks of the eight (8) methods: EOSA, ABC, WOA, BOA, PSO, GA, DE, and HGSO. The test statistics ($\chi^2$) value of 249.847380 was obtained along with degrees of freedom (df) of 7 and significance level (Asymptotic Significance) of 0.001. We discovered a statistically significant difference in the performance of the eight (8) methods compared based on the values of $\chi^2(7) = 249.847380$, $p = 0.001$. The existence of this significant difference then necessitated the need for Wilcoxon signed-rank tests. Each method (optimization algorithm) was uniquely combined with EOSA to determine where the significance lies. Running the test, the results in Table 13 shows the post hoc output of the Wilcoxon signed-rank tests. The post hoc analysis confirms that there was a statistically significant reduction in perceived effort in the ABC-EOSA ($Z = -5.602$, $p = 0.001$), WOA-EOSA ($Z = -$

3.635, $p = 0.001$), BOA-EOSA ($Z = -4.277$, $p = 0.001$), PSO-EOSA ($Z = --3.532$, $p = 0.001$), DE-EOSA ($Z = --5.613$, $p = 0.001$), GA-EOSA ($Z = --5.64$, $p = 0.001$), and HGSO-EOSA ($Z = -5.415$, $p = 0.001$).

**Table 13:** Wilcoxon Post hoc test of EOSA with each of the selected optimization methods

|  | ABC - EOSA | WOA - EOSA | BOA - EOSA | PSO - EOSA | DE - EOSA | GA - EOSA | HGSO - EOSA |
|---|---|---|---|---|---|---|---|
| Z | -5.602[b] | -3.635[b] | -4.277[b] | -3.532[b] | -5.613[b] | -5.645[b] | -5.415[b] |
| Asymp. Sig. (2-tailed) | .000 | .000 | .000 | .000 | .000 | .000 | .000 |
| a. Wilcoxon Signed Ranks Test ||||||||
| b. Based on negative ranks. ||||||||

In summary, the argument that based on the outcome of the exhaustive experimentation done in this study, EOSA has shown to be a search algorithm capable of finding better solutions in a tight competition with state-of-the-art optimization algorithms. Also, that the proposed algorithm demonstrated that it could find far better solutions with fewer computational requirements compared with ABC, WOA, BOA, PSO, GA, DE, and HGSO methods.

### 6. Conclusion

This paper has presented a novel optimization algorithm, EOSA, based on the propagation model of the deadly Ebola virus and its associated disease. The study has shown how the bio-inspired algorithm derived its efficiency from the dynamic mechanism of moving individuals in the population through the susceptible, infected, quarantined, hospitalized, recovered, and dead sub-population. The study presented an improved version of the propagation model of the Ebola virus disease, which was further translated into a mathematical model. The resulting model was applied to the design of the design the novel optimization algorithm EOSA. We have applied EOSA to two sets of benchmark functions consisting of forty-seven (47) classical and over forty-four (44) constrained IEEE CEC-benchmark functions. The outcome of extensive experimentation to determine the algorithm's performance showed that it provides performance on a par with other population-based methods. Although the EOSA metaheuristic algorithm did not show superior performance in all cases, a significant outcome confirms it is very potent in handling optimization problems.

Moreover, considering the no-free lunch theorem, we safely conclude that the optimization fits into the body of recognized and viable optimization algorithms in the literature. A more interesting outcome of the proposed algorithm is the computational demand required for its performance. The result from the experimentation showed that the CPU time for the completion of the algorithm was substantially lower than some state-of-the-art optimization algorithm. This advantage will be more relevant and pronounced in applying the algorithm to real-world optimization problems, emphasizing time management. As future work, this study intends to investigate further strategies capable of maximizing greater balance between the exploration and exploitation phase of the algorithm. Also, the constraint of the new algorithm might be overcome using a hybridization solution with another optimization algorithm, demonstrating characteristics of eliminating the constraint.

**Declaration of Competing Interest**

The authors declare that they have no known competing financial interests or personal relationships that could have appeared to influence the work reported in this paper.